\documentclass[journal]{IEEEtran}

\usepackage{amsmath,amsfonts,amssymb}
\usepackage{algorithmic}
\usepackage{algorithm}
\usepackage{array}
\usepackage{textcomp}
\usepackage{stfloats}
\usepackage{url}
\usepackage{verbatim}
\usepackage{graphicx}
\usepackage{cite}
\usepackage{color} 

\ifCLASSOPTIONcompsoc
    \usepackage[caption=false,font=normalsize,labelfont=sf,textfont=sf]{subfig}
\else
    \usepackage[caption=false,font=footnotesize]{subfig}
\fi

\usepackage{textcomp}
\usepackage{xcolor}
\usepackage{tipa}
\usepackage{tabularx}
\usepackage{multirow}
\usepackage{pdfpages}

\def\BibTeX{{\rm B\kern-.05em{\sc i\kern-.025em b}\kern-.08em
    T\kern-.1667em\lower.7ex\hbox{E}\kern-.125emX}}
    
\begin{document}

\title{
DynamicLip: Shape-Independent Continuous Authentication via Lip Articulator Dynamics
}

\author{Huashan Chen, Yifan Xu, Yue Feng, Ming Jian, Feng Liu, Pengfei Hu, \\ Kebin Peng, Sen He and Zi Wang
\thanks{Huashan Chen and Ming Jian are with Institute of Information Engineering, Chinese Academy of Sciences, Beijing 100085, China (email: chenhuashan@iie.ac.cn; jianming@iie.ac.cn).}
\thanks{Yifan Xu, Yue Feng and Feng liu are with Institute of Information Engineering, Chinese Academy of Sciences, Beijing 100085, China, and also with School of Cyber Security, University of Chinese Academy of Sciences, Beijing 101408, China (email: xuyifan@iie.ac.cn; fengyue2023@iie.ac.cn; liufeng@iie.ac.cn).}
\thanks{Pengfei Hu is with School of Computer Science and Technology,
Shandong University, Qingdao 266237, China (email: phu@sdu.edu.cn).}
\thanks{Kebin Peng is with Department of Computer Science, East Carolina University, Greenville 27858, USA (email: 
 pengk24@ecu.edu).}
\thanks{Sen He is with Department of Systems and Industrial Engineering, University of Arizona, Tucson 85718, USA (email: 
 senhe@arizona.edu).}
\thanks{Zi Wang is with Department of Computer \& Cyber Sciences, Augusta University, Augusta 30912, USA (email: zwang1@augusta.edu).}
\thanks{Corresponding author: Zi Wang.}
}
\IEEEpubid{0000--0000/00\$00.00~\copyright~2024 IEEE}

\maketitle

\begin{abstract}
Biometrics authentication has become increasingly popular due to its security and convenience; however, traditional biometrics are becoming less desirable in scenarios such as new mobile devices, Virtual Reality, and Smart Vehicles. For example, while face authentication is widely used, it suffers from significant privacy concerns. The collection of complete facial data makes it less desirable for privacy-sensitive applications. 
Lip authentication, on the other hand, has emerged as a promising biometrics method. 
However, existing lip-based authentication methods heavily depend on static lip shape when the mouth is closed, which can be less robust due to lip shape dynamic motion and can barely work when the user is speaking.
In this paper, we revisit the nature of lip biometrics and extract shape-independent features from the lips. We study the dynamic characteristics of lip biometrics based on articulator motion. Building on the knowledge, we propose a system for shape-independent continuous authentication via lip articulator dynamics. This system enables robust, shape-independent and continuous authentication, making it particularly suitable for scenarios with high security and privacy requirements.
We conducted comprehensive experiments in different environments and attack scenarios and collected a dataset of 50 subjects. The results indicate that our system achieves an overall accuracy of 99.06\% and demonstrates robustness under advanced mimic attacks and AI deepfake attacks, making it a viable solution for continuous biometric authentication in various applications.
\end{abstract}


\section{Introduction}

\IEEEPARstart{B}{iometric} authentication has become increasingly popular due to its ability to provide secure and convenient access control. According to a recent study released in 2023, the global biometric authentication market was valued at USD 34.27 billion by 2022, growing at a compound annual growth rate (CAGR) of 20.4\% from 2023 to 2030\cite{biometrics_market}. The adoption of biometric systems is driven by their applications across various sectors, including banking, healthcare, and mobile devices. For instance, Apple reported that its Face ID technology is used by millions of users worldwide, with over 30,000 infrared dots mapped onto a user’s face for high security\cite{face_id_apple}. Additionally, a survey found that 77\% of consumers are satisfied using biometrics on their smartphone, highlighting the growing consumer acceptance\cite{mitek_biometrics}. This widespread adoption underscores the increasing reliance on biometric authentication for enhancing both security and user convenience.

However, traditional biometric methods like facial, voice, and fingerprint recognition are increasingly inadequate for new scenarios in mobile devices, VR, and Smart Vehicles. Facial authentication raises privacy concerns due to comprehensive data collection. Voice authentication is prone to background noise, mimicry, and replay attacks, compromising security and privacy in public spaces. Fingerprint authentication requires physical contact, raising hygiene issues. Moreover, these methods often lack continuous authentication, crucial for dynamic scenarios like VR and Smart Vehicles. Thus, there is a growing need for non-intrusive biometric solutions that provide continuous authentication and better address privacy concerns.

\IEEEpubidadjcol

\begin{figure}[!t]
    \centering
    \includegraphics[width=0.49\textwidth]{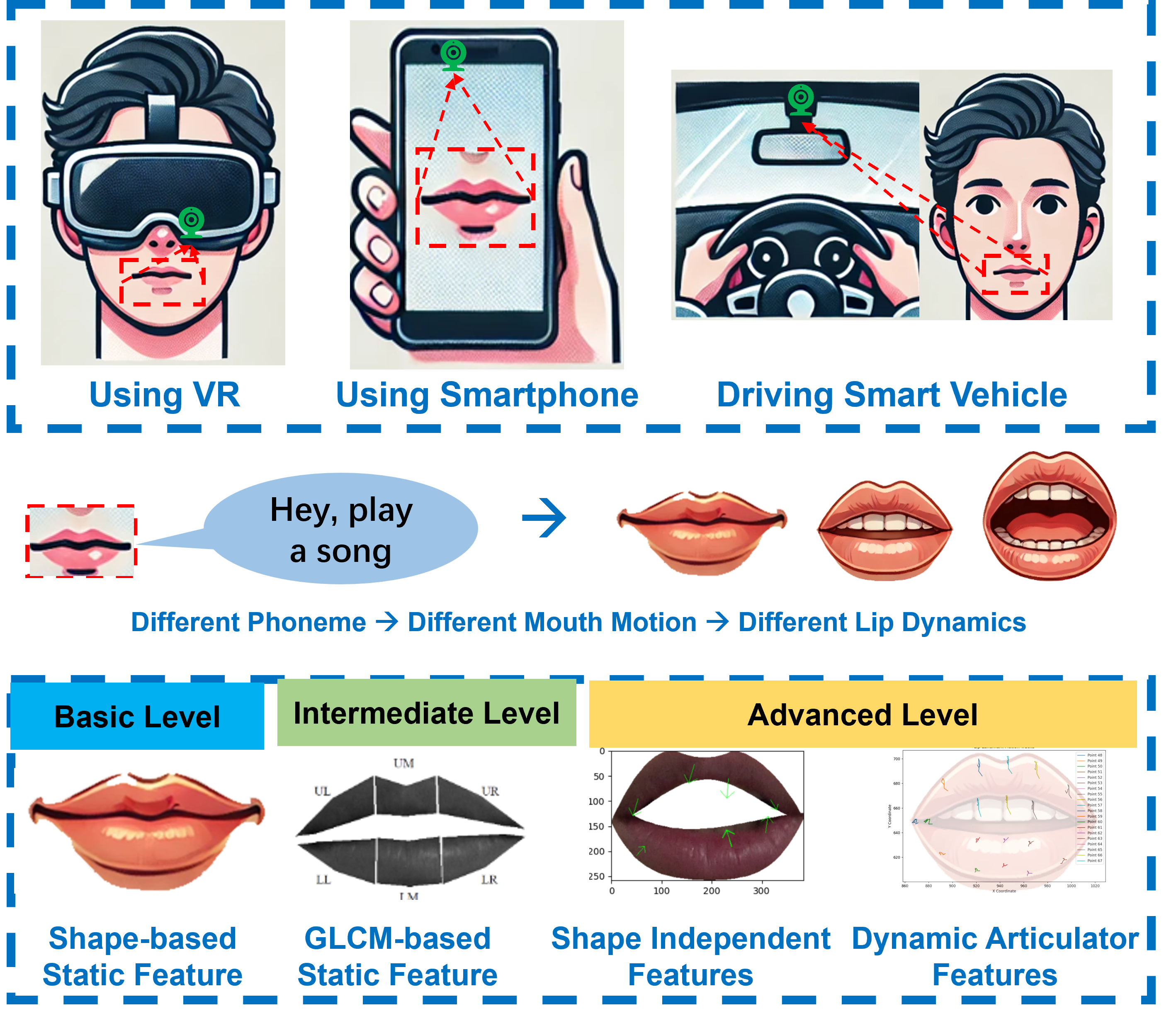}
    \caption{Core idea of the proposed system.}
    % \vspace{-0.3cm}
    \label{fig:core_idea}
    % \vspace{-0.5cm}
\end{figure}

Lip biometrics has been a good alternative compared with traditional methods. 
Prior works have explored lip biometric-based user authentication using static images of lips~\cite{farrukh2023lip,smacki2010lip,wrobel2013method,jain2018cheiloscopic}. 
However, these existing works heavily rely on static lip shapes, limiting their robustness and reliability. Static methods can be less effective due to variations in lip shapes from different expressions or speaking styles, making them susceptible to spoofing and less adaptable to dynamic authentication needs.
Some recent works have explored lip dynamics for user authentication, each with its own set of limitations. For instance, Simultaneous Localization and Mapping (SALM) serves as an additional layer of authentication alongside a password and does not offer the benefit of continuous authentication \cite{yuan2017salm}. SilentKey utilizes ultrasonic signals to detect motion, which can be sensitive to environmental noise, potentially exposing it to side-channel attacks \cite{tan2018silentkey}. LipAuth focuses on smiling motion, which limits its effectiveness in scenarios requiring dynamic lip movements, such as speaking \cite{kuang2023lipauth}.
Also, these approaches mainly extract features from lip videos without fully considering the underlying articulatory motion and the nature of dynamic lip movements. Consequently, they may not capture the complete complexity and variability of lip dynamics, which can impact the accuracy and robustness of biometric authentication in dynamic scenarios. 

In this paper, we propose a novel biometric of dynamic lip articulator motion for shape-independent and continuous authentication.
To address the challenges in biometric authentication, we present a system that leverages the dynamic and articulatory nature of lip movements to ensure robust and continuous user verification.
In particular, we first revisit the nature of lip biometrics and extract the shape-independent features from lip patterns. This approach provides robustness against variations in lip shape, making it more reliable and more secure compared to static lip-based authentication systems. By focusing on the dynamic aspects of lip movements, we mitigate the issues related to static shape variations caused by factors such as speaking and cosmetic changes.
Secondly, we study the phonetics and lip movement. By analyzing the dynamic motion of the lips during speech, our system categorizes lip motion into three major types. 
This categorization allows our system to provide continuous authentication under various speaking conditions, ensuring ongoing user verification without interrupting user activities. This feature is particularly beneficial in scenarios such as VR and Smart Vehicles, where seamless and continuous authentication is critical, as shown in \figurename \ref{fig:core_idea}.
We then build a lip feature hierarchy that captures both the global and local characteristics of lip movements and shapes. This hierarchical model enables a detailed and comprehensive representation of the dynamic lip articulator motion, enhancing the system's ability to differentiate among users accurately.
For authentication, we propose a Siamese neural network model that effectively authenticates the user's unique lip dynamic features. It enhances the reliability of our authentication system by combining multiple weak classifiers to form a strong classifier, thereby improving overall performance.
We conducted comprehensive experiments in various environments and collected a dataset of 50 subjects. Our results indicate that our system robustly achieves high accuracy, making it a viable solution for continuous biometric authentication in diverse applications.

Our system offers significant advantages over traditional biometric methods, primarily in terms of security, privacy, usability, and implementation. It enhances security by reducing the risk of spoofing due to the dynamic nature of lip movements, making it resilient to mimic attacks, advanced mimic attacks, and AI deepfake attacks. In terms of privacy, our system focuses solely on lip movements, avoiding the extensive collection of facial data and thus minimizing the risk of unauthorized surveillance, identity theft, and data breaches. Usability is improved with a non-contact and hands-free operation, addressing hygiene concerns and making it suitable for public or shared devices, as well as scenarios like VR and Smart Vehicles. The system adapts to changes in lip shape, ensuring consistent authentication and enhancing security in real-time applications. From an implementation perspective, our system is low-cost and can be easily deployed using standard smartphone cameras, making it accessible for widespread use. \textcolor{black}{The contributions of our paper include:}

\begin{itemize}
    \item We revisit the nature of lip biometrics and understand the phonetics and the articulatory characteristics of lip movements, enabling advanced feature extraction. 
    \item We build a lip feature hierarchy that captures the basic, intermediate, and advanced levels of characteristics of dynamic lip movements. 
    \item Based on this knowledge, we design a robust system for shape-independent continuous authentication. Our system ensures non-intrusive and hand-free authentication.
    \item We collect a comprehensive dataset involving 50 subjects speaking carefully designed phrases that contain commonly used phonemes and associated lip movements, providing a representative sample of everyday speaking scenarios. The dataset is uniquely structured to capture natural lip dynamics during speech while also preserving both lip and facial information, making it a valuable resource for the research community to study both lip-specific and full-face authentication methods. 
    \item Our experimental results using this dataset demonstrate an accuracy of 99.06\% in different everyday environments and scenarios, validating our system's effectiveness.
\end{itemize}

The remainder of the paper is organized as follows. 
Section \ref{sec:preliminary} presents the preliminary concepts and foundations necessary for our approach. 
Section \ref{sec:system_design} describes the system design. 
In Section \ref{sec:experiment_setup}, we detail the experimental setup and data collection process. 
Section \ref{sec:evaluation} presents the evaluation results of our system. 
In Section \ref{sec:related_work}, we review related work.
Section \ref{sec:conclusion} concludes the paper and outlines directions for future work.

\section{Preliminary}
\label{sec:preliminary}

\subsection{Static Lip Biometrics}
Traditional lip biometrics have been widely used in identity recognition due to the uniqueness, permanence, and stability of lip print. Researchers found that the lip print of all subjects did not show the same pattern \cite{tsuchihashi1974studies}, indicating their potential as biometric features. Lips contain rich information characterized by their unique patterns, making them important biological features for identity recognition \cite{zhou2023lip}.

\subsection{Traditional Lip Features}

Traditional lip features include color and shape.
Color is an important feature that allows humans to quickly distinguish between objects and images. The lip color features can be obtained by calculating statistical measures in the color space \cite{choras2010lip}. 
Also, people have similarities in general shapes of lip, in-depth research demonstrates that the variations in lip shapes among different people are more significant than the changes in lip shape within photos of the same individual \cite{gomez2002biometric}. 
Lip shapes have uniqueness, such as lip width, upper/lower lip height, lip perimeter, lip region area, and the ratio of related parameters. 
The shape features of lips can be analyzed using commonly used shape analysis descriptors in image processing, such as central moments, Zernike moments, Hu moments, Malinowska ratio, Feret ratio, Blair-Bliss ratio, Danielsson ratio, Haralick ratio, Lp1 ratio, and Lp2 ratio \cite{choras2007human}. 
However, due to the variability of lip shape, using lip shape for identity authentication has limitations. 
For example, differences in lip shape among individuals during various expressions, especially when speaking, make relying solely on static lip shape less desirable and effective. The dynamic nature of lip movements during speech necessitates combining shape features with other stable features to ensure reliable biometric authentication.
In our proposed work, we focus on extracting more reliable features, i.e., shape-independent dynamic articulator lip motion, which are less impacted by external factors and more consistent across different conditions. This helps in achieving more consistent and accurate biometric authentication, even under different speaking conditions.

\subsection{Lip Texture Dynamics Beyond Shape}
Lip textures are the unique and complex arrangements of wrinkles and grooves on the outermost layer of the lips. These textures are effective for personal identification due to their distinctiveness and permanence. Suzuki and Tsuchihashi \cite{suzuki1971new} categorized lip grooves into six types: a clear-cut groove running vertically across the lip, a partial-length groove of type I, a branched groove, an intersected groove, a reticular groove, and others. Each individual's lip texture pattern remains unchanged throughout their lifetime, making lip texture a reliable biometric feature.
The advantages of using lip textures for biometric authentication lie in their inherent uniqueness and robustness. Lip texture features are less sensitive to lighting conditions and whether the subject wears makeup, compared to color and shape features. The complex and unique nature of lip textures, combined with minimal variation across different expressions and health conditions, provides higher robustness for identification.

In our proposed work, to address the challenges of texture feature extraction, especially in dynamic scenarios, we utilize advanced image processing and signal processing techniques to ensure accurate and reliable extraction of lip texture features. By tracking the motion and distortion of the lip patterns, we leverage these dynamic characteristics for shape-independent biometric authentication.

\subsection{Articulator Lip Motion and Phonetics}
Phoneme, the smallest phonetic unit in a language, forms the basis of word pronunciation and is categorized into vowels and consonants. Peter Ladefoged et al. \cite{ladefoged2006course} define a vowel as any unobstructed sound occurring in the middle of a syllable, making vowels essential for word formation. According to the International Phonetic Association, vowel pronunciation involves specific mouth and tongue movements, with lip openings classified as close, close-mid, open-mid, and open. Individuals exhibit unique mouth-opening habits, and even the same person varies the size of their mouth opening when pronouncing different vowels.
Therefore, we divide vowels into three categories based on the size of their mouth at the time of pronunciation: big, middle, and small. 
By incorporating this classification into the speaker authentication process, individuals' speaking patterns can be matched with their own baseline measurement results, thereby improving the efficiency of authentication.
In our proposed work, to address the challenges of dynamic lip movement and phonetics, we develop a robust classification system that categorizes lip movements based on phoneme articulation. This system enhances the accuracy of biometric authentication by accounting for variations in speaking habits and improving the system's adaptability to different phonetic contexts.

\begin{figure*}[!ht]
    \centering
    \includegraphics[width=0.88\textwidth]{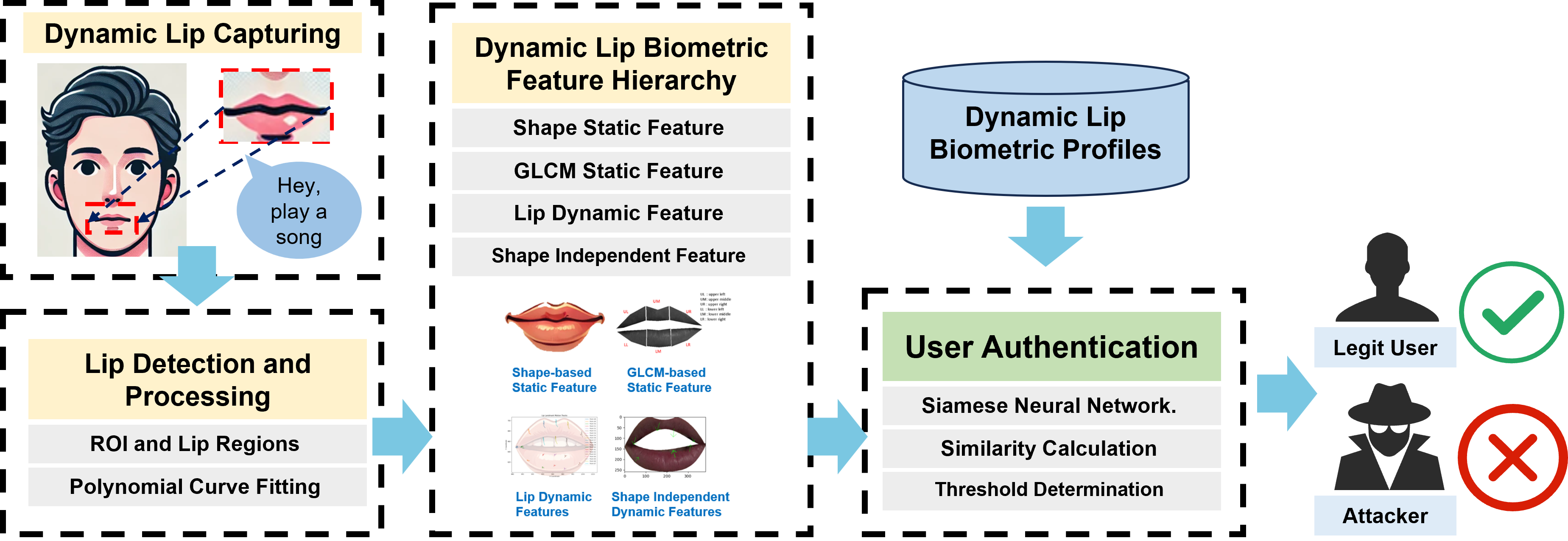}
    \caption{System flow of the proposed system.}
    \label{fig:sys_flow}
    % \vspace{-0.5cm}
\end{figure*}

\section{System Design}
\label{sec:system_design}

The core idea underlying our system is to leverage the dynamic and articulatory nature of lip movements for robust, shape-independent, and continuous biometric authentication. 
By capturing and analyzing both static and dynamic lip features through our comprehensive feature hierarchy, we ensure high-accuracy authentication regardless of the user's speaking state, making it particularly suitable for real-world applications requiring hands-free and seamless user verification. 
Unlike traditional approaches that only work in restricted conditions, our system maintains consistent performance whether the user is in a natural resting state or actively speaking. 
Our proposed system for dynamic lip biometric authentication comprises four main stages: Dynamic Lip Capturing, Lip Detection and Processing, Dynamic Lip Biometric Feature Extraction, and User Authentication. The system workflow is illustrated in Fig. \ref{fig:sys_flow}.

Firstly, in the Dynamic Lip Capturing stage, the system captures the dynamic lip motion of the user while the user is speaking. This is particularly useful in scenarios such as VR and Smart Vehicles where hands-free and continuous authentication is required. 
Next, the captured lip motion is preprocessed in the Lip Detection and Processing stage. Lip detection and processing are performed using techniques such as dlib for detecting lip landmarks and polynomial curve fitting to accurately map the lip contours. This step ensures that the dynamic lip movements are accurately captured and prepared for feature extraction.
Also, in the Dynamic Lip Biometric Feature Extraction stage, we build a dynamic hierarchy based on our knowledge of the nature of shape-independent lip patterns and articulator motion phonetics. 
Finally, in the User Authentication stage, we authenticate the user based on their dynamic lip biometric profile. Using a siamese network model, we perform similarity calculations and threshold determination to distinguish between legitimate users and attackers. 

\subsection{Attack Model}
\label{sec:Attack Model} 

We designed and considered three different attack scenarios to evaluate the robustness of our biometric system: mimic attack, advanced mimic attack, 
and AI deepfake attack. These scenarios cover a spectrum of potential threats to ensure our system can maintain high security in real-world applications.

In the mimic attack scenario, attackers have limited knowledge of the system’s operation or its features. They attempt to bypass the system by mimicking the \textcolor{black}{lip motion of a} legitimate user speaking the same passphrase. 
In the advanced mimic attack scenario, an attacker uses a static photograph of the legitimate user's face or lips, held in front of their own face, to deceive the system. This evaluates the system's capability to differentiate between live, dynamic lip movements and static images, given that static features might be mimicked to some extent using a photograph, but dynamic lip movements during speech are much harder to replicate with a static image. 
In addition, for AI deepfake attack scenario, an attacker uses AI-generated deepfake technology to replace the lip movements of the legitimate user in a video. This scenario assesses the system's ability to detect and resist sophisticated AI-generated forgeries. Despite the advances in deepfake technology, it cannot fully reproduce the unique lip patterns and dynamics required by our system, thus failing to bypass the authentication process.

The effectiveness of our system against these attacks is discussed in the later evaluation section (Section \ref{sec:eva_attackanalysis}), demonstrating the robustness and reliability of our dynamic, shape-independent lip biometric features in various scenarios.

\subsection{Lip Detection and Processing}
\label{sec:Lip Detection} 

\begin{figure}[t]
    \centering
    \includegraphics[width=0.6\linewidth]{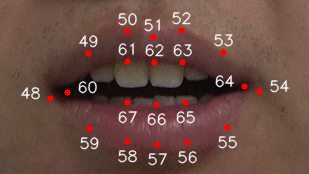}
    \caption{The lip region formed by 20 landmarks (point 48 to 67).}
    \label{fig:20_landmarks}
\end{figure}

The challenge in dynamic lip detection is accurately capturing the varying shapes and movements of the lips, especially during speech. Our approach addresses this by first detecting facial landmarks using an ensemble of regression trees, mapping 68 (x, y) coordinates to facial features such as the eyes, nose, mouth, and chin \cite{Kazemi2014One}. The region of interest (ROI) is the mouth area.
We focus on the coordinates for the mouth, specifically points 48 to 67 for a total of 20 landmarks, as shown in Figure \ref{fig:20_landmarks}. To describe the lip shape more accurately, we use a multi-segment polynomial curve fitting method to form the complete lip contour by connecting piece-wise curves. This extracted lip contour is then used to create a mask that isolates the lip area. This design enables precise dynamic lip detection, providing a robust foundation for further feature extraction. By accurately modeling the lip contour, our system enhances its ability to capture and analyze lip movements during speech, which is critical for reliable biometric authentication.

\subsection{Dynamic Lip Biometric Feature Hierarchy}

\begin{figure}[!t]
    \centering
    \includegraphics[width=0.49\textwidth]{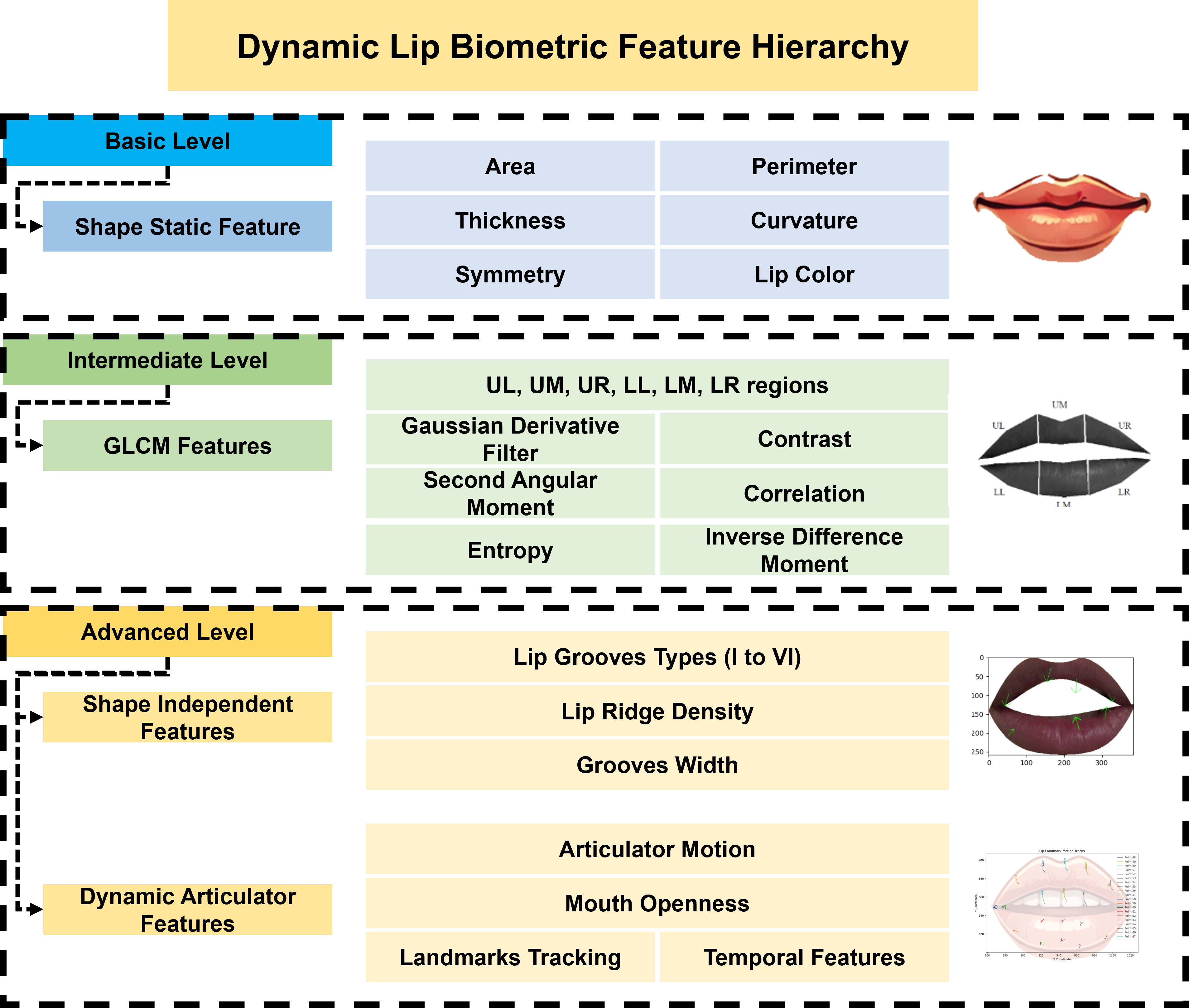}
    \caption{Dynamic lip biometric feature hierarchy.}
    \label{fig:feature_hierarchy}
    % \vspace{-0.5cm}
\end{figure}

Based on the preliminary knowledge of lip biometric, we then design the Dynamic Lip Biometric Feature Hierarchy, which includes three levels: Basic, Intermediate, and Advanced, as shown in Fig. \ref{fig:feature_hierarchy}. The \textbf{Basic Level} encompasses fundamental Shape Static Features, including area, perimeter, thickness, curvature, symmetry, and lip color, providing essential geometric and colorimetric data. The \textbf{Intermediate Level} involves Gray-Level Co-occurrence Matrix (GLCM) Features, focusing on texture analysis through regions like UL, UM, UR, LL, LM, and LR, alongside the application of Gaussian Derivative Filters to extract intricate texture details. The \textbf{Advanced Level} integrates more complex analyses with Shape Independent Features and Lip Dynamic Articulator Features. Shape Independent Features include lip groove types, ridge density, and groove depth, which are critical for identifying unique lip patterns. Lip Dynamic Features encompass articulator motion, mouth openness, landmarks tracking, and temporal features, capturing the dynamic aspects of lip movement and changes over time. This hierarchical approach ensures a comprehensive and detailed analysis of lip biometrics for robust biometric identification.

\subsection{Basic Level}

\subsubsection{Shape-based Static Feature}
\begin{figure}[!t]
    \centering
    \subfloat[]{
        \includegraphics[width=0.2\textwidth]{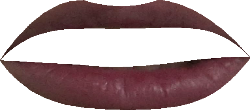}
    }
    \hspace{0.01\textwidth}
    \subfloat[]{
        \includegraphics[width=0.2\textwidth]{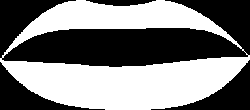}
    }
    \caption{Region of Interest (ROI). (a) Cropped and scaled ROI. (b) Barbarized ROI.}
    \label{fig:roi}
\end{figure}

To capture the basic geometric characteristics of the lips when the mouth is static, we first extract Shape-based Static Features. After obtaining the Region of Interest (ROI), we perform further shape feature extraction to obtain the static features of the lip, which includes area, perimeter, and thickness. To ensure data consistency and completeness, we resize and binarize the images to 250×110 pixels as shown in Fig. \ref{fig:roi}, where the ROI is white and the background is black. To get the area of the lip, we count the total number of white pixels in the binarized image, which represents the covered area of the lip region. Second, we use the contour detection algorithm in OpenCV to obtain the lip contour, and calculate the perimeter of the contour to obtain the lip perimeters. Finally, we calculated the thickness of the upper and lower lip. Specifically, we scan the binarized image column by column and calculate the continuous segment length of the white pixels in each column to obtain the thickness.

\subsection{Intermediate Level}
\subsubsection{GLCM-based Feature}

\begin{figure}[!t]
    \centering
    \includegraphics[width=0.35\textwidth]{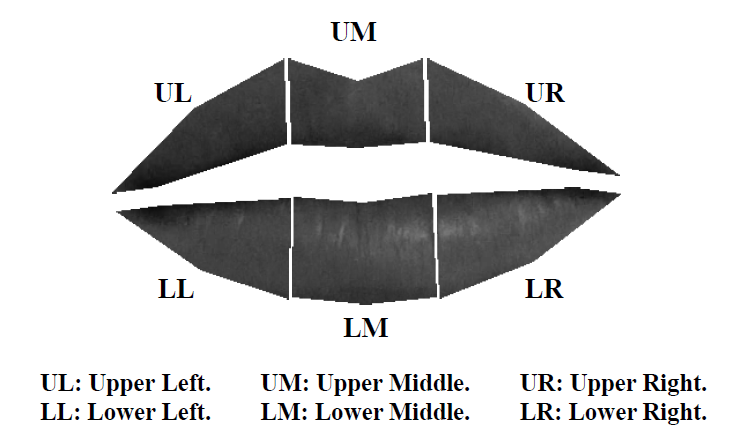}
    \caption{Six regions of the lip.}
    \label{fig:6regions}
    % \vspace{-0.5cm}
\end{figure}

To extract the features of the image in more detail, we divide the lip region into six parts\cite{choras2011lip}: UL, UM, UR, LL, LM, LR, as shown in Fig. \ref {fig:6regions}. Then we extract features for each region.

\begin{figure}[!t]
    \centering
    \includegraphics[width=0.48\textwidth]{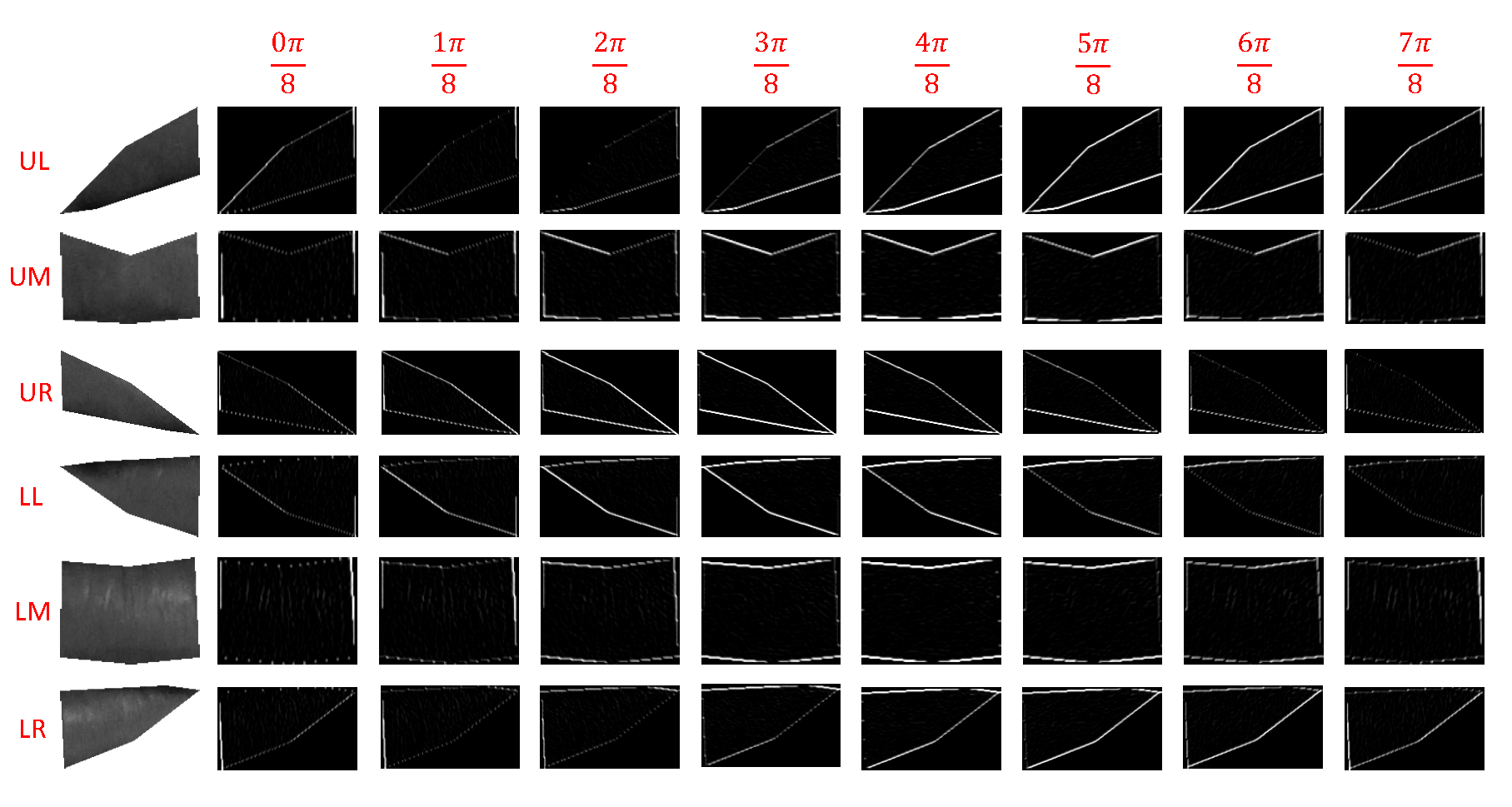}
    % \vspace{-0.3cm}
    \caption{Steerable responses of six regions.}
    \label{fig:after_filter}
    % \vspace{-0.3cm}
\end{figure}

In this paper, we use the second derivative of a Gaussian function as a filter kernel, and analyze the texture features of images from 8 directions.
The 8 results of the six regions after Steerable Filters are shown in Fig. \ref{fig:after_filter}.

The calculation of the features is based on the GLCM \cite{mohanaiah2013image} of the image after the filtering operation. The GLCM is a statistical method of examining texture that considers the spatial relationship of pixels. The features derived from the GLCM are used to quantify various aspects of the texture, providing insights into the image's structure and patterns. 
There are five features used includes Second Angular Moment (ASM), Contrast, Correlation, Inverse Difference Moment, and Entropy.
Thus, for one frame in a video, we obtain six sets of 5×8 feature matrices. 
% Given the size of these matrices, they are somewhat large for subsequent classification.
To capture the essence of the features and reduce the complexity, we employ the Principal Components Analysis (PCA) method. PCA helps to downscale the features by extracting the main information from the data. Consequently, the 8 directions are finally reduced to two dimensions, significantly simplifying the classification process while retaining the most critical aspects of the lip movement patterns.

\subsection{Advanced Level}

\subsubsection{Lip Shape Independent Features}

\begin{figure}[!t]
    \centering
    \subfloat[]{\includegraphics[width=0.2\textwidth]{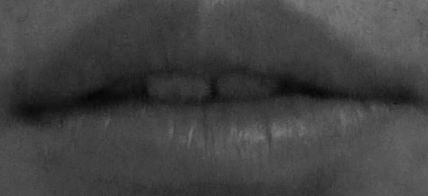}\label{Grayscale}}
    \hspace{0.01\textwidth}
    \subfloat[]{\includegraphics[width=0.2\textwidth]{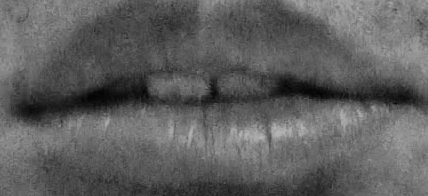}\label{CLAHE}} \\
    % \hspace{0.01\textwidth}
    \subfloat[]{\includegraphics[width=0.2\textwidth]{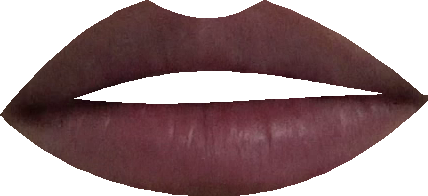}\label{Lip-only}}
    \hspace{0.01\textwidth}
    \subfloat[]{\includegraphics[width=0.2\textwidth]{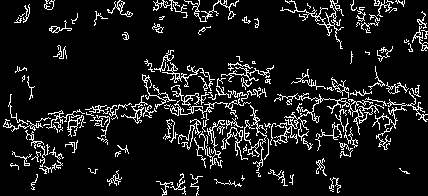}\label{Canny}} \\
    \subfloat[]{\includegraphics[width=0.2\textwidth]{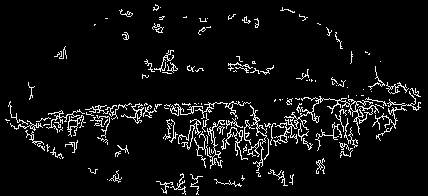}\label{Bilateral Filter}}
    \hspace{0.01\textwidth}
    \subfloat[]{\includegraphics[width=0.2\textwidth]{figs/feature/HT/HT3_bilateralFilter.png}\label{Keep only lip}}
    % \vspace{-0.1cm}
    \caption{Pre-processing operations to extract lip shape independent features. (a) Grayscale. (b) CLAHE. (c) Lip-only. (d) Canny. (e) Bilateral filter (f) Keep only lip.}
    \label{Pre-processing}
    % \vspace{-0.5cm}
\end{figure}

Lip Shape Independent Features provide a different perspective on lip biometrics by focusing on lip pattern characteristics that are not directly related to lip shape.
To effectively extract dynamic lip patterns, we first convert the image to grayscale and apply the Contrast Limited Adaptive Histogram Equalization (CLAHE) to enhance local contrast and improve image details, as shown in Fig. \ref{Grayscale} and Fig. \ref{CLAHE}. Next, we perform edge detection using the Canny algorithm to identify clear boundaries and a bilateral filter is then applied to remove noise while preserving edge information, as illustrated in Fig. \ref{Canny} and Fig. \ref{Bilateral Filter}. Following these steps, we fit the lip contours using the method described in Section Lip Detection, as depicted in Fig. \ref{Lip-only}. Finally, we remove non-lip regions in the filtered image to obtain the final result shown in Fig. \ref{Keep only lip}.

\begin{figure}[!t]
    \centering
    \subfloat[]{\includegraphics[width=0.2\textwidth]{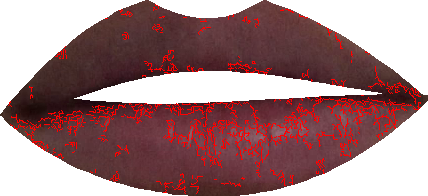}\label{HT6_seg}}
    % \vspace{0.01\textwidth}
    \hspace{0.01\textwidth}
    \subfloat[]{\includegraphics[width=0.2\textwidth]{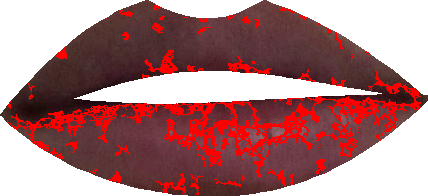}\label{seg_link2}}
    \vspace{0.01\textwidth}
    \\
    \subfloat[]{\includegraphics[width=0.2\textwidth]{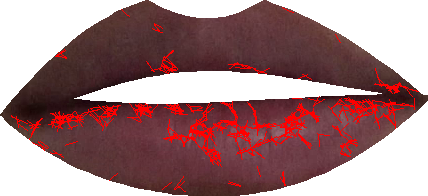}\label{length}}
    \hspace{0.01\textwidth}
    \subfloat[]{\includegraphics[width=0.2\textwidth]{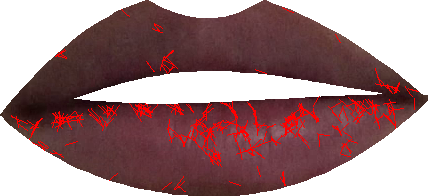}\label{angle}}
    % \hspace{0.01\textwidth}
    % \vspace{-0.5cm}
    \caption{Lip print detection. (a) Segmentation. (b) Connection. (c) Length filter. (d) Angle filter.}
    \label{HT}
    % \vspace{-0.5cm}
\end{figure}

Next, we compare the preprocessed image in Fig. \ref{Keep only lip} to retain parts of the straight lines that are white pixels, as shown in Fig. \ref{HT6_seg}. To address the large number of short segments that appear after screening, which are not conducive to subsequent feature calculations, we connect line segments that are less than 2 pixels apart, as shown in Fig. \ref{seg_link2}. Considering the actual condition of lip prints, we retain segments longer than 10 pixels to avoid noise, as shown in Fig. \ref{length}. Horizontal lines, likely representing the outer contour of the lips, are filtered to keep lines with angles between 40-140°. Ultimately, we identify the texture lines of lips in Fig. \ref{angle}, which will be used to calculate subsequent dynamic features.

\begin{figure}[!t]
    \centering
    \subfloat[]{\includegraphics[width=0.2\textwidth]{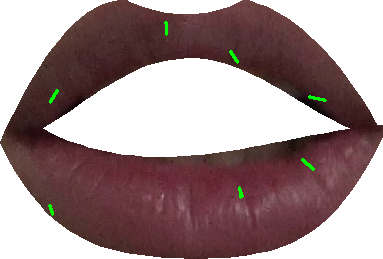}\label{frame1}}
    % \vspace{0.01\textwidth}
    \hspace{0.01\textwidth}
    \subfloat[] {\raisebox{.20\height}{\includegraphics[width=0.2\textwidth]{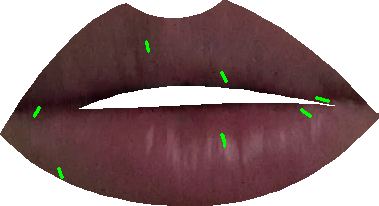}}\label{frame2}}
    % \vspace{0.01\textwidth}
    \\
    \subfloat[]{\includegraphics[width=0.2\textwidth]{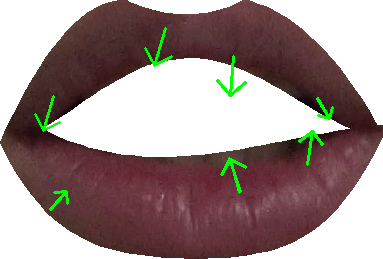}\label{Motion}}
    \hspace{0.01\textwidth}
    \subfloat[]{\raisebox{.3\height}{\includegraphics[width=0.2\textwidth]{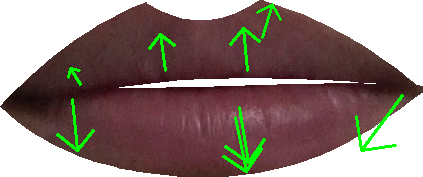}}\label{large2}}
    % \vspace{0.01\textwidth}
    \caption{Motion vectors of two frames. (a) Status 1. (b) Status 2. (c) Close vector. (d) Open vector.}
    \label{Motion_vectors}
    % \vspace{-0.5cm}
\end{figure}

To capture shape-independent dynamic features, we calculate changes in lip texture between two frames. Given the randomness of lines detected after the Hough Transform (HT), we match lines at the same positions in two frames by dividing the lip into six regions (Fig. \ref{fig:6regions}). In each of the six regions, we compare texture lines in two frames, selecting matched lines based on minimal movement, length difference, and angle. We use the center coordinates of these matched lines to compute motion vectors, focusing on the longest vectors from each region. This results in a set of eight motion vectors for two frames. As shown in Fig. \ref{Motion_vectors}, by comparing different mouth statuses, we calculate the motion vectors as green vectors. The green motion vectors show the direction and distance of the lip movements. Finally, we compute statistical characteristics such as Average Motion Vector, Average Motion Magnitude, Average Motion Direction, Trajectory Length, and Trajectory Curvature as dynamic texture features.

\subsubsection{Dynamic Articulator Features
}
\label{sec:DynamicArticulatorFeature}
\begin{figure*}[!t]
\centering
    \subfloat[]{
        \includegraphics[width=0.245\textwidth]{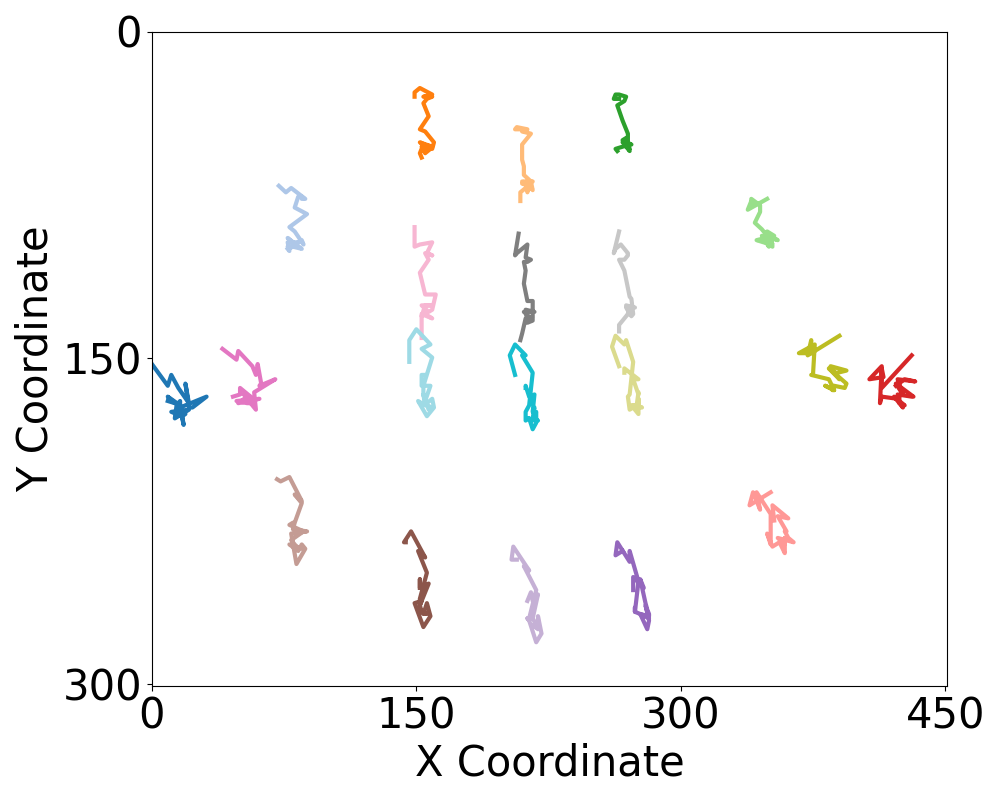}
        \label{fig:big_corr}
    }
    \subfloat[]{
        \includegraphics[width=0.245\textwidth]{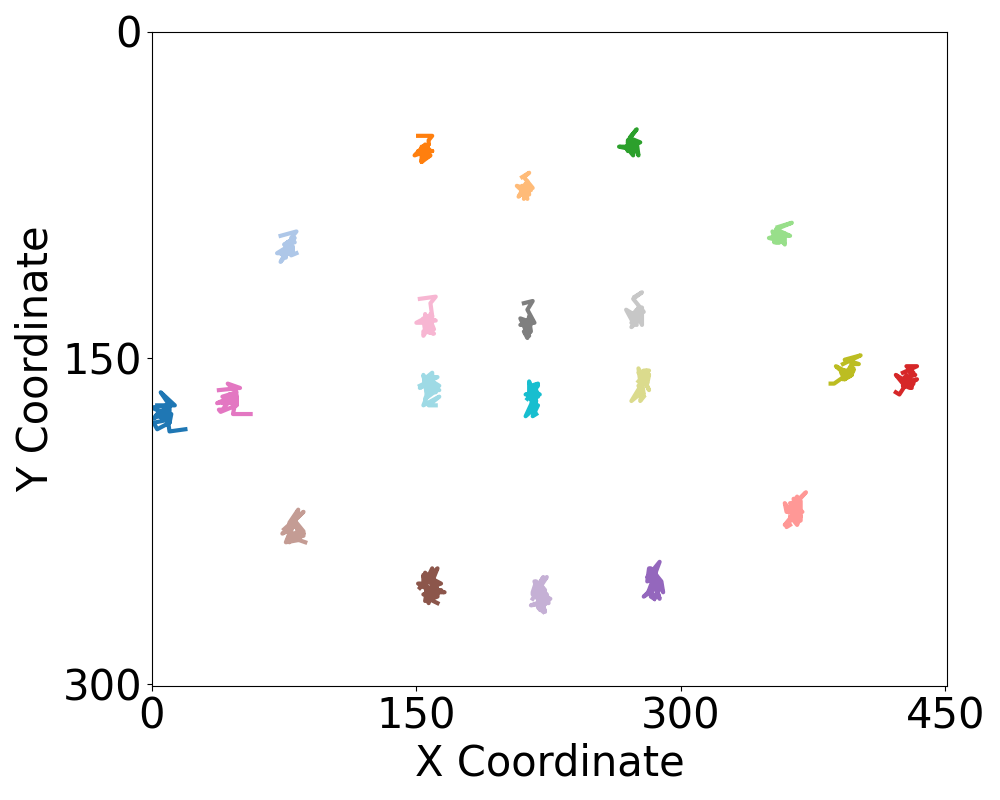}
        \label{fig:middle_corr}
    }
    \subfloat[]{
        \includegraphics[width=0.245\textwidth]{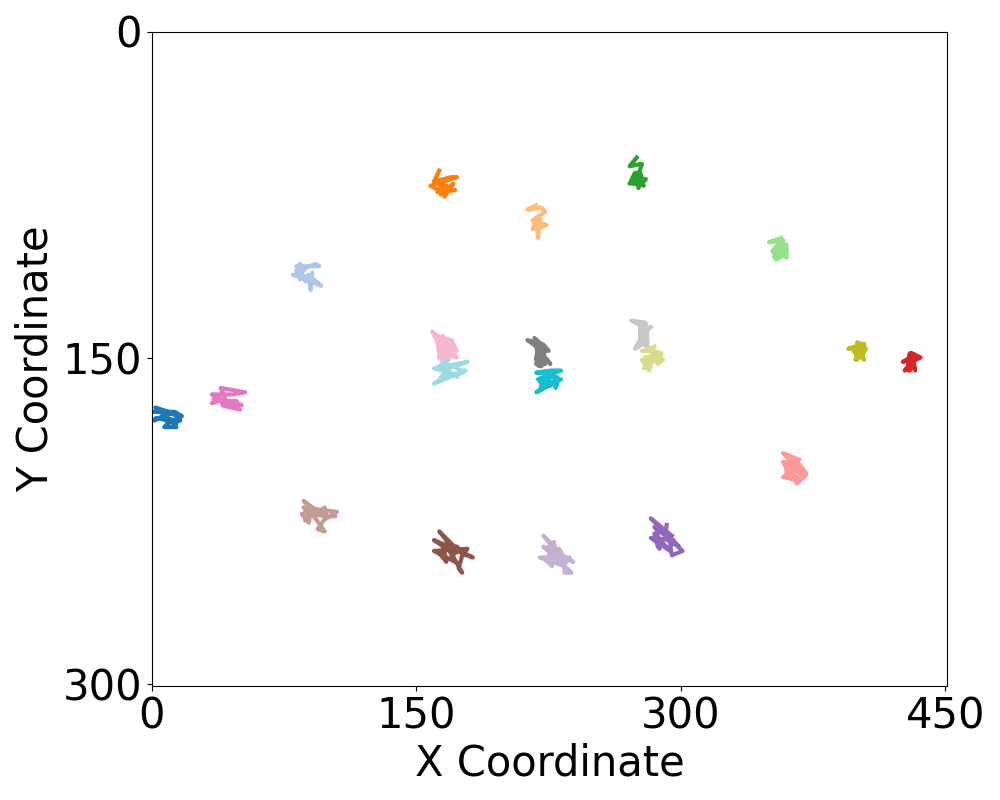}
        \label{fig:small_corr}
    }
    \subfloat[]{
        \includegraphics[width=0.245\textwidth]{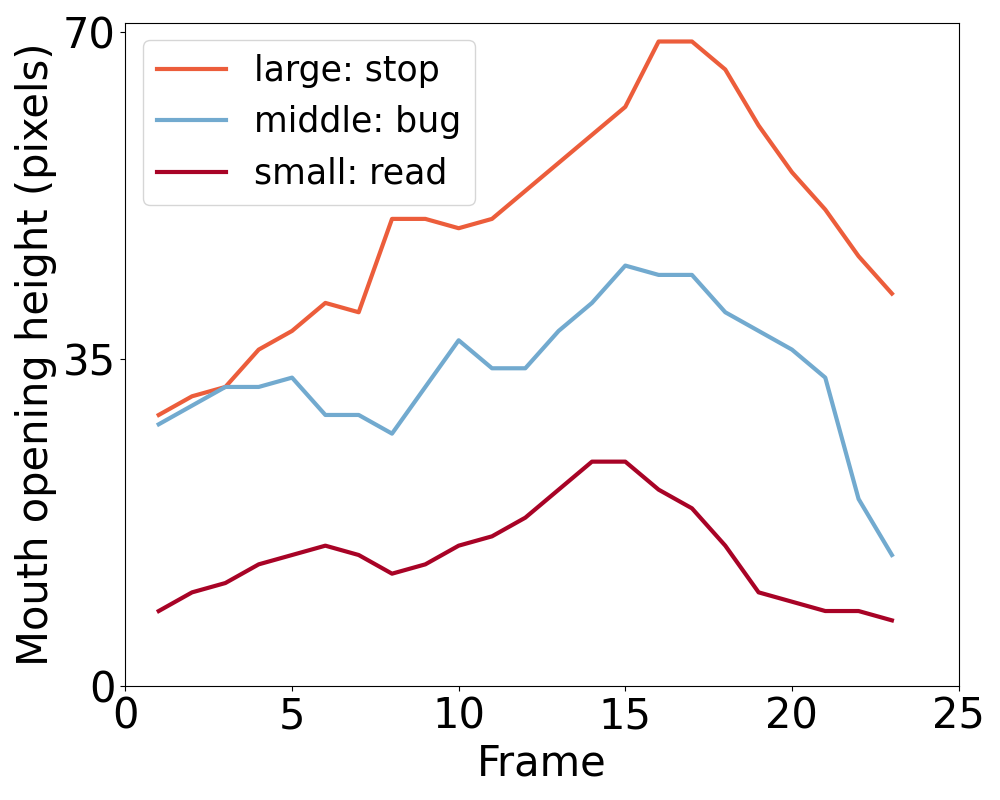}
        \label{fig:mouth_open_change}
    }
    \caption{Longitudinal movement trajectories of the lips while saying different words. (a) Big opening: word ``stop". (b) Middle opening: word ``bug". (c) Small opening: word ``read". (d) Mouth height.}
    \label{fig:correlation}
    % \vspace{-0.5cm}
\end{figure*}

% (Phonetics)
Dynamic Articulator Features capture the motion and changes in the lips over time, offering insights into the dynamics of lip movement. 
Based on the theory of phonetics, when people speak different vowels, their mouths exhibit different shapes and levels of openness. We conducted a preliminary study on mouth shape changes when users speak commonly used words in daily life or IoT environments, such as interacting with AI assistants. As shown in Fig. \ref{fig:correlation}, Fig. \ref{fig:big_corr} to Fig. \ref{fig:small_corr} display the movement trajectories of the landmarks when different people speak various words in different categories. Fig. \ref{fig:mouth_open_change} demonstrates mouth openness can be roughly divided into three levels.
From our analysis of commonly spoken words, we discovered that mouth shapes can be categorized into three distinct categories: large mouth opening, medium mouth opening, and small mouth opening, i.e.,
small mouth opening: /\textipa{I}/, /\textipa{i:}/, /\textipa{I@}/, /\textipa{U}/, /\textipa{u:}/, /\textipa{U@}/, 
medium mouth opening: /\textipa{\ae}/, /\textipa{e}/, /\textipa{@}/, /\textipa{3:}/, /\textipa{2}/, /\textipa{eI}/, /\textipa{OI}/, /\textipa{@U}/, /\textipa{e@}/, and
large mouth opening: /\textipa{O:}/, /\textipa{6}/, /\textipa{A:}/, /\textipa{aI}/, /\textipa{aU}/.

Then, by analyzing the movement trajectories of the lip key points, we found that these three categories correspond to three unique profiles of each user when speaking. We also found that lip movement characteristics in each profile vary between different individuals when speaking the same phoneme. This categorization allows us to extract temporal features of lip movement, providing a more accurate description of the dynamic behavior of the lips and enhancing the robustness and accuracy of our biometric authentication system~\cite{Li2008Novel}.

To illustrate the dynamic articulator feature, we construct a sequence of 20 landmarks. These 20 landmarks, shown in Fig. \ref{fig:correlation}, are points 48 to 67 mentioned in Section \ref{sec:Lip Detection}. For each landmark, we record its x-coordinate and y-coordinate in each frame to form a motion matrix.
To represent the synergistic changes in lip movements, we calculated the correlation coefficients between lip landmarks movement trajectories:
\vspace{0.1cm}
\begin{equation}
\vspace{0.1cm}
\label{eq:correlation}
r_{i,j}=\frac{\sum_{k=1}^n(x_i^k-\overline{x_i})(x_j^k-\overline{x_j})}{\sqrt{\sum_{k=1}^n(x_i^k-\overline{x_i})^2}\sqrt{\sum_{k=1}^n(x_j^k-\overline{x_j})^2}}
\end{equation}
where $r_{i,j}$ denotes to the correlation coefficient between the $i^{th}$ landmarks and the $j^{th}$ landmarks, $\overline{x_i}$ denotes to the average of the coordinates of the $i^{th}$ landmarks, and $n$ denotes to the total number of frames. 
With the motion matrix and \eqref{eq:correlation}, we can construct the correlation coefficient symmetric matrix of x-coordinate and y-coordinate.

\subsection{Classifier Training for Authentication}
\begin{figure}[!t]
    \centering
    \includegraphics[width=\linewidth]{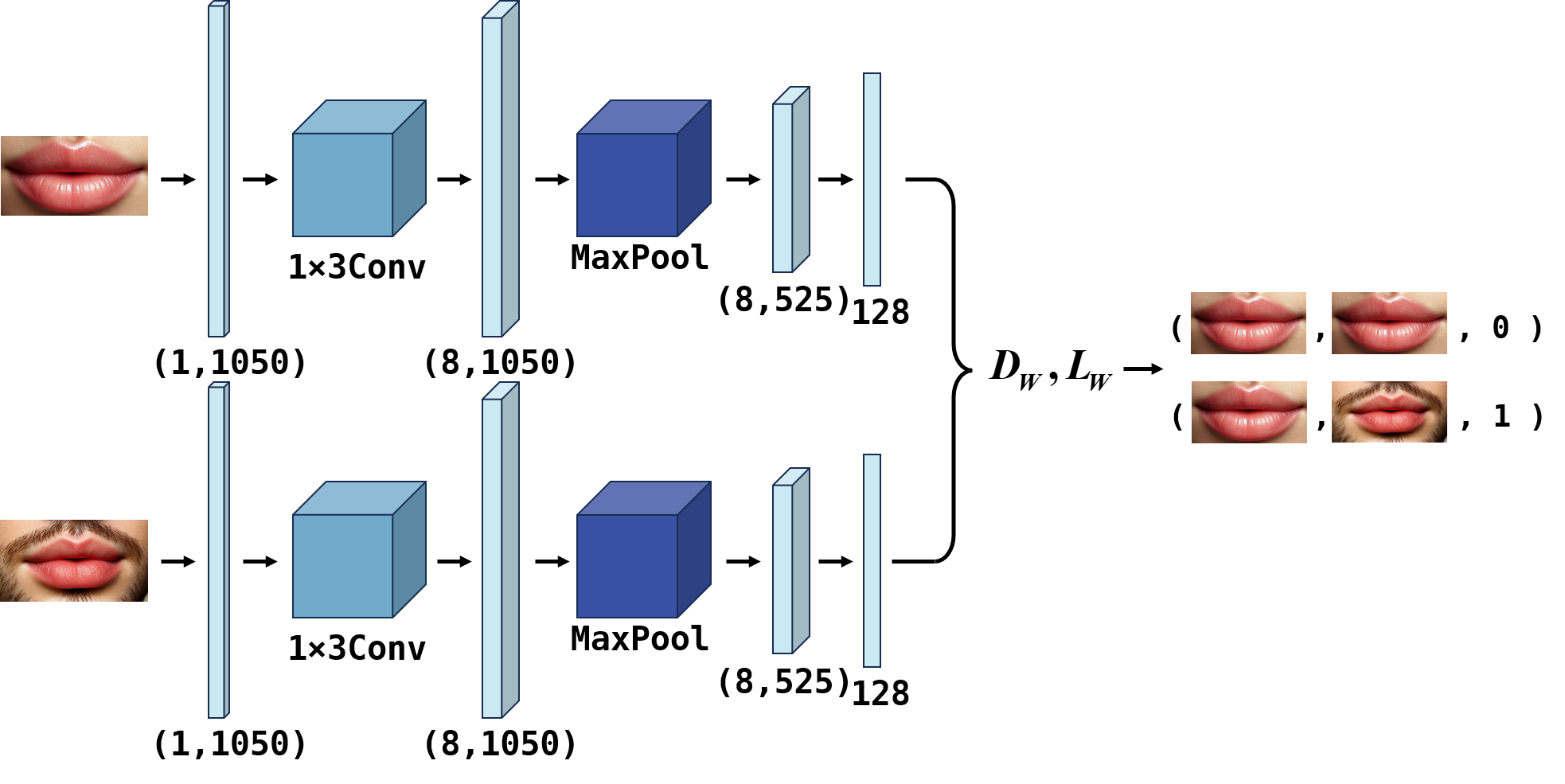}    \caption{Siamese network architecture.}
    \label{fig:siamese}
    % \vspace{-0.3cm}
\end{figure}

After obtaining the features of each participant, our system proceeds to the authentication process. We employed a Siamese Network, which consists of two identical neural networks. The network outputs two feature vectors, and we calculate the Euclidean distance between these vectors to measure their similarity, as shown in Fig. \ref{fig:siamese}. The model is trained using contrastive loss, enhancing its ability to distinguish between positive and negative pairs. 

\section{Experiment Setup}
\label{sec:experiment_setup}

\subsection{Dataset and Environment}

The selection of a suitable dataset is crucial for the following experiments. To better evaluate our system and analyze various factors such as articulator motion, viewing angles, lighting conditions, and makeup effects, we recorded our own dataset for dynamic lip authentication (https://github.com/xxxxx)\footnote{Upon acceptance, we will publish our datasets.}.

\subsubsection{Articulatory content design}

Based on our knowledge and analysis of mouth articulator motion, we have categorized the mouth articulator motion into three different categories. We then classified the vowel phonemes into these categories based on phonetics, as discussed in Section \ref{sec:DynamicArticulatorFeature}.

We design the passphrases with two key considerations in mind. First, we select phrases commonly used in daily life, particularly focusing on voice assistant commands in IoT environments, ensuring real-world applicability. Second, we carefully choose phrases that cover all phoneme categories identified in our lip dynamics study, encompassing the full range of lip motions from small to large mouth openings. This comprehensive selection ensures our system can effectively authenticate users across the diverse spectrum of natural lip movements that occur during regular speech. Table~\ref{sentences} presents these carefully selected passphrases.

\begin{table}[t]
% \centering
\begin{center}
\caption{Common IoT instructions adopted in the evaluation.}
\vspace{-0.5cm}
\label{sentences}
    \begin{tabular}{|>{\centering\arraybackslash}p{0.225\textwidth}|>{\centering\arraybackslash}p{0.225\textwidth}|}
        \hline
        Hey Siri, turn on the light. & Cook dinner. \\
        \hline
        Hey Google, play a song. & Charge my phone. \\
        \hline
        Find my keys. & Play a video. \\
        \hline
        Turn off the fan. & Turn on the clock. \\
        \hline
        Hi, turn on the fan. & Turn off the TV. \\
        \hline
        Show a video. & Lock the door. \\
        \hline
        Start the car. & Turn off the light. \\
        \hline
        Read my messages. & Start a fire. \\
        \hline
        Read this page. & Book a plane. \\
        \hline
        Open the door. & Set an alarm. \\
        \hline
        Close the lock. & Send my phone. \\
        \hline
        Turn on the alarm. & Turn on the music. \\
        \hline
        Check my bag. & Take a note. \\
        \hline
        Send a text. & Make a list. \\
        \hline
        Show the book. & Wash the cup. \\
        \hline
    \end{tabular}
\end{center}
\vspace{-0.2cm}
\end{table}

\subsubsection{Volunteer recruitment}
According to the experimental design, we prepare a volunteer recruitment notice and ask volunteers to sign an informed consent form, ensuring they are aware of data usage and the confidentiality of their personal information. The recruitment message is disseminated through social media platforms. We recruit 50 volunteers (22 females and 28 males), all of whom participated in the regular experimental setup. Additionally, 14 of these volunteers took part in an additional impact study. The regular experimental recording lasted approximately 20–30 minutes, while the additional impact study lasted about 2 hours. Volunteers are compensated with \$40 for the regular recording and \$100 for participating in the additional impact study.

\subsubsection{Recording environment and process}
We set up six identical experimental environments for simultaneous recording in a closed room with curtains drawn and three lights turned on, ensuring an illuminance of 200 (± 10) lux as measured by a light intensity meter. Volunteers sat in front of a white wall with their heads close to the wall and remained still during recording. 
A SONY FDR-AX60 video camera recorded 4k 25 fps video, along with the front and rear cameras of various smartphones placed on tripods directly in front of the volunteer's head.
The smartphones used included iPhone 11, Redmi K60, HUAWEI nova6, HUAWEI Mate50, and Honor Magic5 Pro for the rear cameras, and HUAWEI Mate50 and Honor Magic5 Pro for the front cameras.
Our designed passphrases were printed and placed under the camera for volunteers to read. Volunteers faced the camera and read each group of passphrases in sequence, repeating them five times.

\subsubsection{Composition of dataset}

\begin{figure*}[!t]
    \centering
    \subfloat[]{\includegraphics[width=0.45\linewidth]{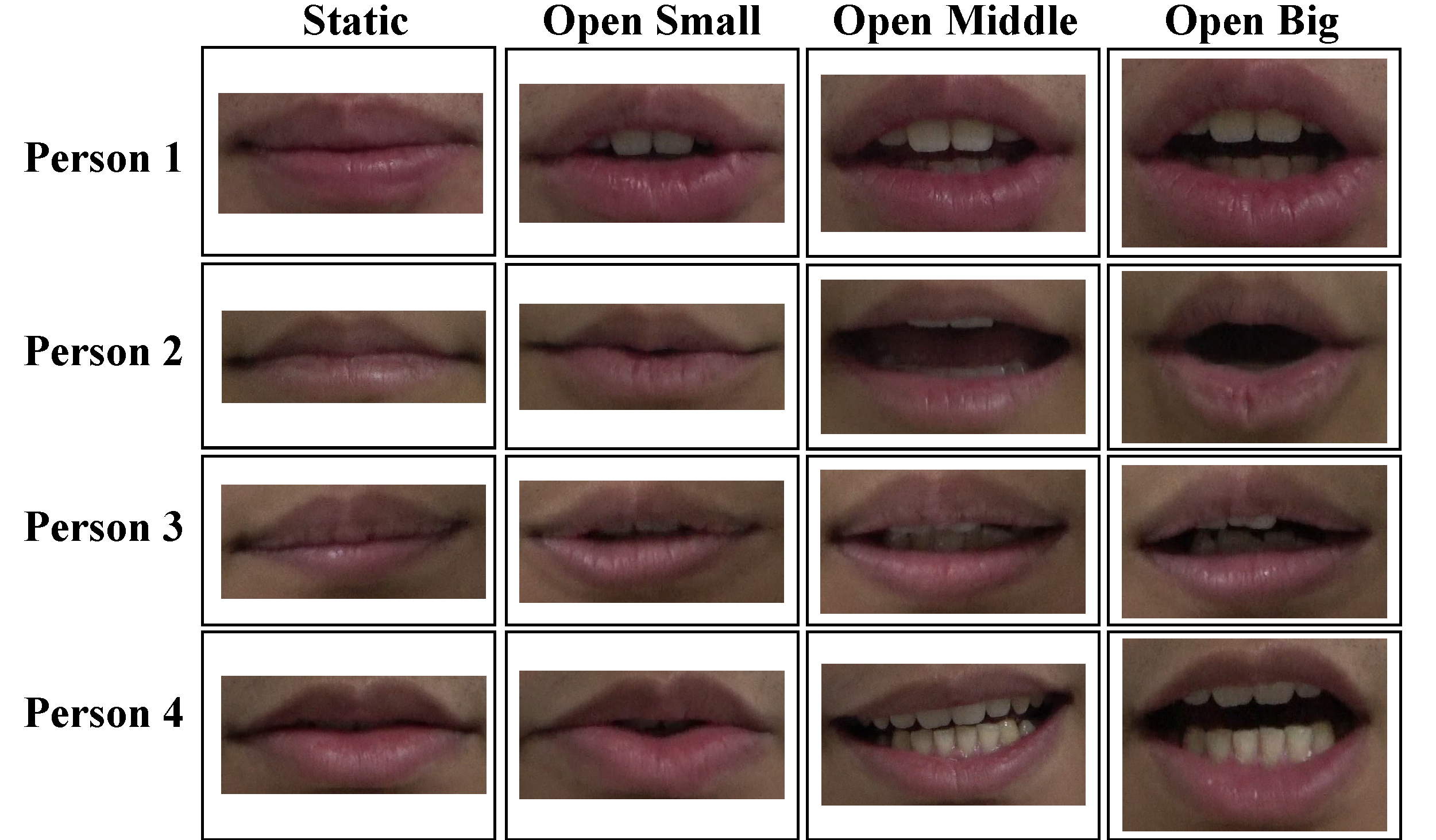}\label{fig:normal_dataset}}
    % \hspace{0.01\textwidth}
   \hspace{0.01\linewidth}
    \subfloat[]{\includegraphics[width=0.45\linewidth]{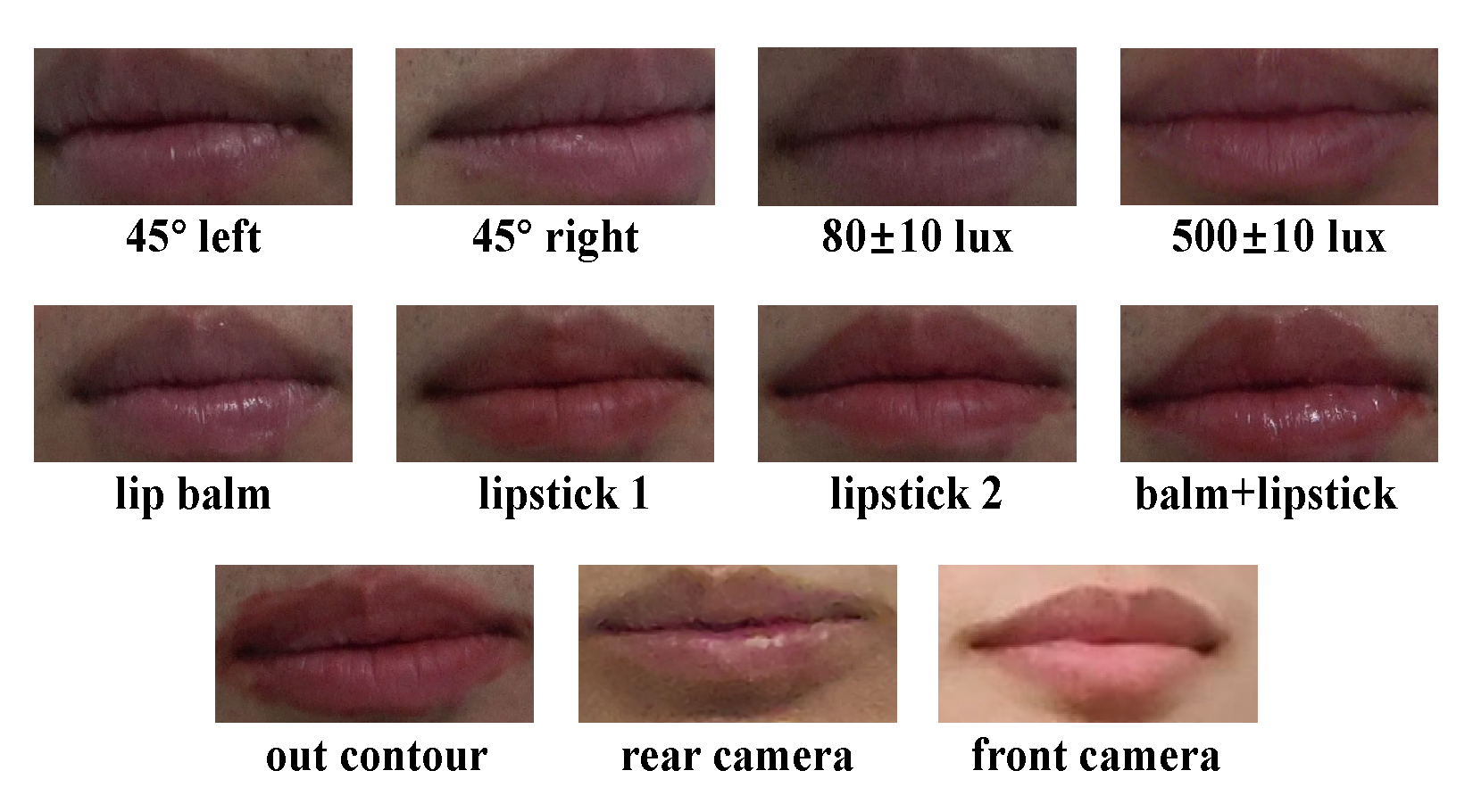}\label{fig:control_dataset}}
    \caption{Dataset Examples. (a) Examples of dataset. (b) Examples of impact study.}
    \label{fig:dataset}
    % \vspace{-0.5cm}
\end{figure*}

The experiment consisted of regular scenarios and additional impact studies. The dataset was collected from 50 volunteers speaking carefully designed passphrases that contain commonly used phonemes and their associated lip movements, providing a representative sample of everyday speaking scenarios. Fig. \ref{fig:normal_dataset} shows the lip regions of four individuals in the dataset in a stationary state as well as in three lip dynamics. The experimental setup for the additional impact study is detailed in Table~\ref{table_setup}. While our analysis focuses on the lip region, the dataset was collected to preserve both lip and facial information, making it a valuable resource for future research in both lip-specific and full-face authentication methods. In addition, the impact studies dataset includes different devices, shooting angles, lip wetness, and colors, as shown in Fig. \ref{fig:control_dataset}. These variations enable comprehensive evaluations of authentication performance across diverse real-world conditions.

\begin{table}[!t]
\begin{center}
\caption{Consider different recording conditions.}
\label{table_setup}
\vspace{-0.5cm}
\resizebox{\linewidth}{!}{
    \begin{tabular}{|m{1.5cm}<{\centering}|m{6cm}<{\centering}|}
    \hline
        \textbf{Factor} & \textbf{Setup} \\ \hline
        \multirow[c]{2}{=}{\centering Recording equipment} & Phone rear camera recording. \\ \cline{2-2}
         & Phone front camera recording. \\ \hline
        \multirow[c]{2}{=}{\centering Angle} & Volunteers turn 45° left relative to the camera. \\ \cline{2-2}
         & Volunteers turn 45° right relative to the camera. \\ \hline
        \multirow[c]{2}{=}{\centering Ambient light} & Turn off the overhead light (80 ± 10 lux). \\ \cline{2-2}
         & Turn on the fill light (500 ± 10 lux). \\ \hline
        \multirow[c]{5}{=}{\centering Lip wetness and color} & Volunteers apply transparent lip balm. \\ \cline{2-2}
         & Apply ``Into you" lipstick. \\ \cline{2-2}
         & Apply ``Mac" lipstick. \\ \cline{2-2}
         & Apply lip balm and ``Mac" lipstick. \\ \cline{2-2}
         & Apply lipstick slightly beyond the lip contour. \\ \hline
    \end{tabular}
    % \end{tabularx}
}
\end{center}
\vspace{-0.2cm}
\end{table}

\subsubsection{Metrics}
To evaluate the authentication performance of our system, we calculated confusion matrix and introduced four different metrics: accuracy, precision, recall, and F1-score. We also utilize the Precision-Recall (PR) curve which represents the performance of the system when the authentication threshold is varying.

\section{Evaluation}
\label{sec:evaluation}

\subsection{Overall Performance}
\begin{figure}[!t]
    \centering
    \includegraphics[width=.9\linewidth]{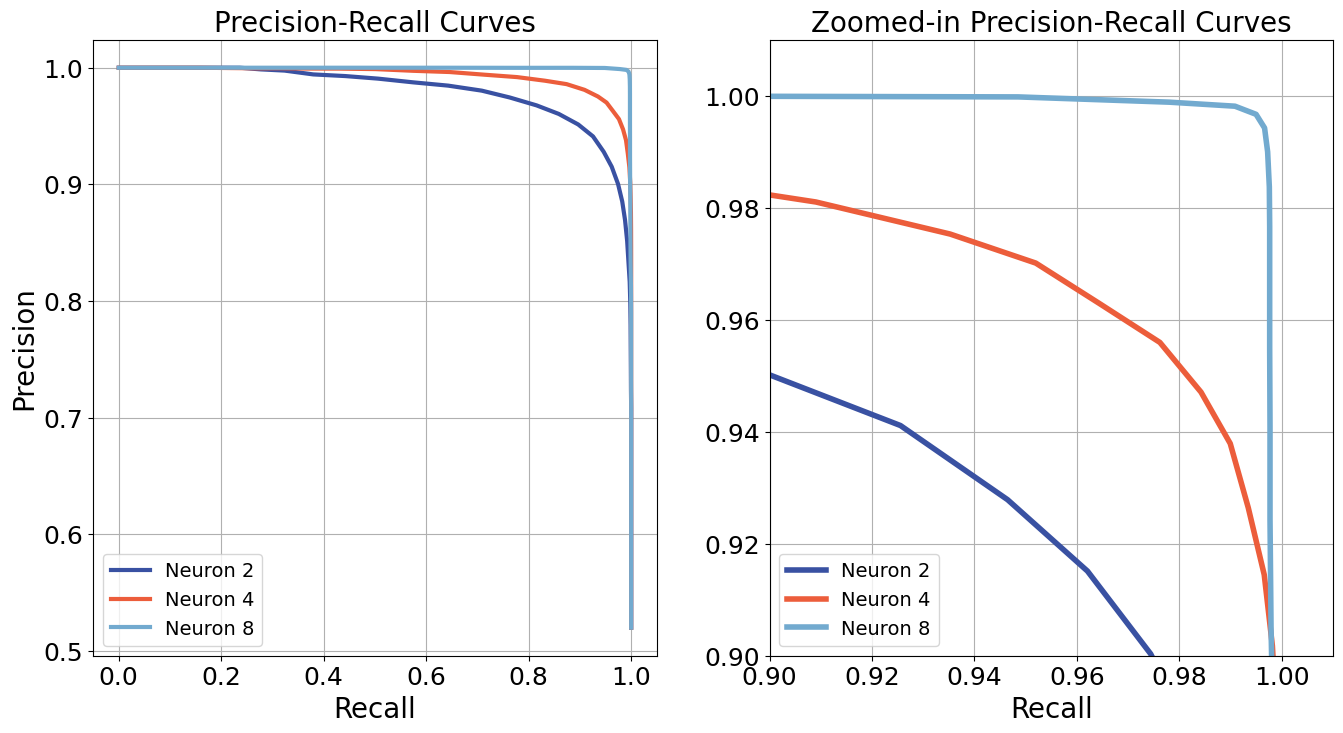}
    \caption{PR curves of different configurations of neurons.}
    \label{fig:neurons_PR_curve}
\end{figure}

We first evaluate our system's performance with all 50 participants. The PR curves for different neuron configurations in the convolutional layer (2, 4, and 8 neurons) are shown in Fig. \ref{fig:neurons_PR_curve}. The 8-neuron configuration provides the best trade-off between discrimination and generalization, so we use it for further experiments. This efficiency is due to our effective feature extraction, eliminating the need for a complex network.

Then, we evaluate our system’s overall performance using tenfold cross-validation, where eightfold is used as the training set, onefold is used as the validation set, and onefold is used as the test set. Table~\ref{tab:overall_performance} summarizes the average and median accuracy, recall, precision, and F-1 score overall. We can observe that DynamicLip can achieve an overall accuracy of  99.06\%, recall of 98.92\%, precision of 99.26\%, and F-1 score of 99.09\% in 10-fold cross-validation. Furthermore, the median accuracy, recall, precision, and F1 score are  99.47\%,  99.59\%,  99.40\%, and  99.49\%, respectively.

\begin{table}[!t]
    \begin{center}
    \caption{Overall Performance.}
    \label{tab:overall_performance}
    \vspace{-0.2cm}
    \begin{tabular}{c|c|c|c}
        \hline 
        & Mean & Median & Deviation\\ 
        \hline
        Accuracy & 99.0588\% & 99.4669\%  & 0.0097 \\ 
        Recall & 98.9273\% & 99.5880\% & 0.0046\\ 
        Precision & 99.2578\%  & 99.3990\% & 0.0046 \\ 
        F-1 Score & 99.0886\% & 99.4879\% & 0.0095\\
        \hline
    \end{tabular}
    \end{center}
    \vspace{-0.2cm}
\end{table}

\subsection{Attack Analysis}
\label{sec:eva_attackanalysis}

\subsubsection{Mimic Attack}
\begin{figure}[!t]
    \centering
    \includegraphics[width=1\linewidth]{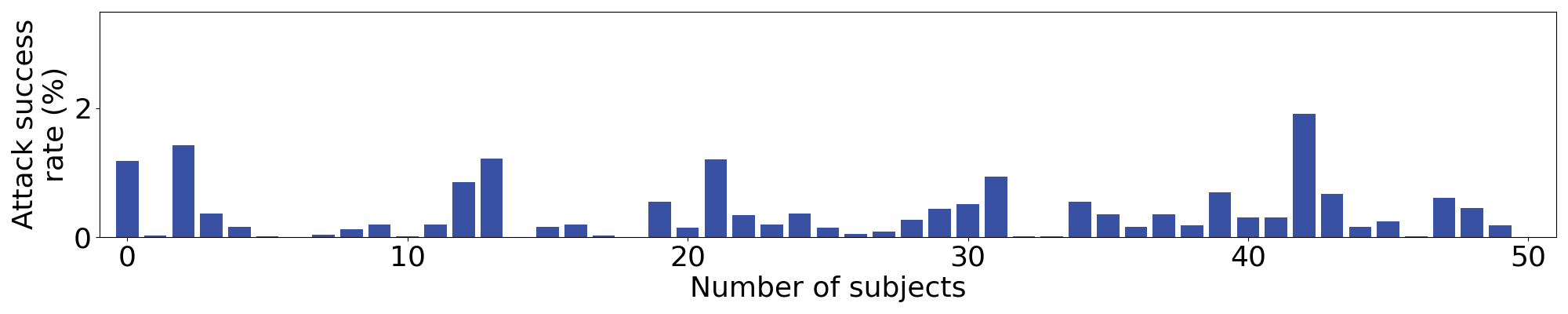} %, height=0.3\linewidth
    \caption{Attack success rate of 50 participants.}
    % \vspace{-0.3cm}
    \label{fig:attack_rate}
    % \vspace{-0.3cm}
\end{figure}

 The attacker has limited knowledge of our system's operation or features, trying to bypass the system using their own lip speaking the same word. During our tests, each of the 50 subjects acts as a legitimate user while the remaining 49 subjects act as attackers, mimicking the lip movements of the legitimate users. As shown in Fig. \ref{fig:attack_rate}, the low average attack success rate of 0.37\% indicates the robustness of our system against mimic attacks. The attackers are unable to successfully bypass the system due to the unique and dynamic nature of the lip biometric features utilized in our authentication process.

\subsubsection{Advanced Mimic Attack}
 The attacker attempts to bypass the system by using a static photograph of the legitimate user's face or lips held in front of their own face. The results indicate that the average attack rate is 0.68\%, showing that while this method might have some success in static scenarios, it fails to mimic the dynamic nature of lip movements. Our system's ability to authenticate users based on dynamic features when they are speaking ensures that a static picture cannot replicate the required dynamic patterns, thereby effectively preventing this type of attack.

\subsubsection{AI Deepfake Attack}
The attackers employed AI-generated deepfake technology to replace the legitimate user's lip movements in a video. Despite the sophistication of deepfake technology, it cannot fully reproduce the unique lip patterns and dynamics required by our system, and the average attack rate is only 1.72\%. This underscores the system's effectiveness in detecting and resisting AI-generated forgeries.

\subsection{Importance of Advance Level Features}

\begin{figure}[!t]
    \centering
    \includegraphics[width=1\linewidth]{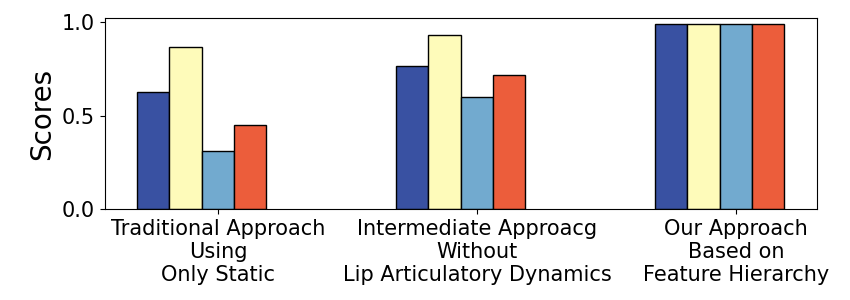}
    \includegraphics[width=0.8\linewidth]{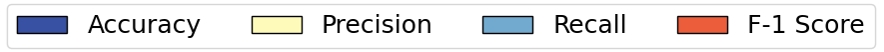}
    \caption{Impact of static and dynamic features.}
    \label{fig:mixed_sta}
\end{figure}

Traditional biometric systems exhibit suboptimal performance by focusing on either static lip features or naive dynamic features. As shown in the first two columns of Fig.~\ref{fig:mixed_sta}, both approaches achieve poor results - the "Traditional Approach Using Only Static" relies solely on static lip shapes, while the "Intermediate Approach Without Lip Articulator Dynamics" uses basic dynamic features without considering comprehensive lip motion characteristics. Neither approach can adequately address real-world scenarios where lip status naturally alternates between static and dynamic states throughout daily activities. These traditional methods only work in restricted conditions - either when lips are completely still or during simple movements - making them impractical for real-life applications where users switch between speaking and non-speaking states.

In contrast, our approach leverages a carefully designed feature hierarchy that effectively handles both static and dynamic lip biometrics. As demonstrated in the third column of Fig.~\ref{fig:mixed_sta}, labeled "Our Approach Based on Feature Hierarchy", the system achieves superior performance across all metrics (Accuracy: 99.06\%, Precision: 99.26\%, Recall: 99.93\%, F1 score: 99.09\%). These results validate our system's effectiveness in both scenarios - when the lips are stationary in their natural state and when they are moving during speech. This robust performance stems from our comprehensive feature hierarchy that captures both static lip patterns and the complex dynamics of lip articulation during speaking, i.e., Lip Dynamic Features and Shape Independent Dynamic Features.

\begin{figure}[!t]
    \centering
    \includegraphics[width=0.6\linewidth]{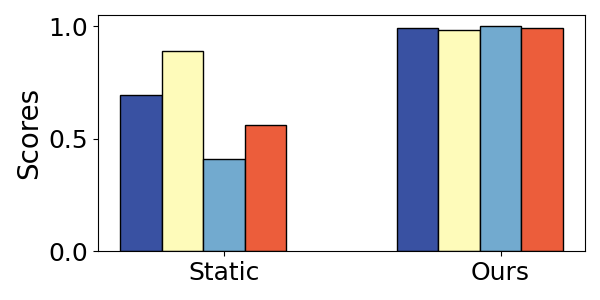}
    \includegraphics[width=0.8\linewidth]{figs/exp/other/legend.png}
    \caption{Evaluation of shape independent feature.}
    \label{fig:outside}
\end{figure}

Similarly, we conducted a study to evaluate the importance of our shape-independent features. Subjects intentionally applied lipstick outside their lip regions to mislead the lip detection features, altering the outline and static shape features. We then compared the performance of systems relying only on shape features to our approach using shape-independent features. The results indicate that traditional approaches perform poorly under these conditions, whereas our approach significantly outperforms them, as shown in Fig. \ref{fig:outside}. This demonstrates the critical importance of shape-independent features in maintaining reliable biometric authentication.

\subsection{Impact of Camera Equipment}
\begin{figure}[!t]
    \centering
    \includegraphics[width=\linewidth]{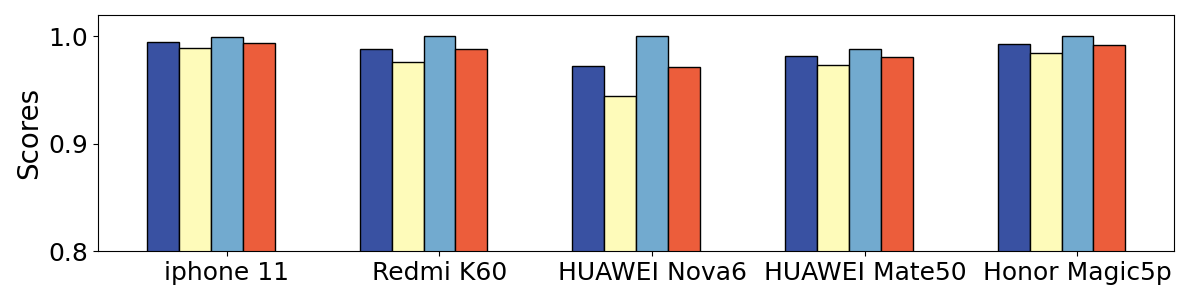}
    \includegraphics[width=0.8\linewidth]{figs/exp/other/legend.png}
    \caption{Evaluations of the impact of phone's rear camera.}
    \label{fig:camera_rear}
\end{figure}

\begin{figure}[!t]
    \centering
    \includegraphics[width=0.6\linewidth]{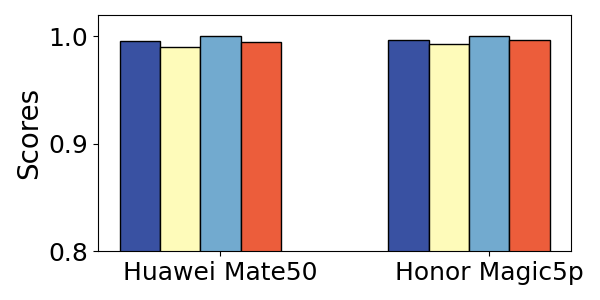}
    \includegraphics[width=0.8\linewidth]{figs/exp/other/legend.png}
    \caption{Evaluations of the impact of phone's front camera.}
    \label{fig:camera_front}
\end{figure}

To evaluate the impact of different camera equipment on our biometric system, we conducted tests using various devices. We used iPhone 11, Redmi K60, HUAWEI nova6, HUAWEI Mate50, and Honor Magic5 Pro for rear cameras, and HUAWEI Mate50 and Honor Magic5 Pro for front cameras.
As shown in Fig. \ref{fig:camera_rear} and Fig. \ref{fig:camera_front}, while the choice of equipment influenced image quality, our system maintained consistent performance across all devices. Despite variations in resolution and detail from different cameras, neither setup significantly impacted the accuracy of our biometric system. This demonstrates that our system is adaptable to various recording equipment, ensuring reliable performance regardless of the device used.


\subsection{Impact of Viewing Angle}
\begin{figure}[!t]
    \centering
    \includegraphics[width=0.6\linewidth]{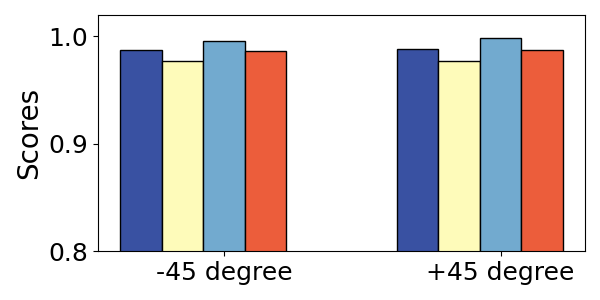}
    \includegraphics[width=0.8\linewidth]{figs/exp/other/legend.png}
    \caption{Evaluation of the impact of viewing angle.}
    \label{fig:angle}
\end{figure}

To evaluate the effect of viewing angle on our biometric system, we conducted tests with 14 subjects turning 45° left and 45° right relative to the camera, as shown in Fig. \ref{fig:control_dataset}.
As illustrated in Fig. \ref{fig:angle}, our system maintained effectiveness despite angle changes. While turning 45° introduced some distortion and partial occlusion, it had minimal impact on recognition accuracy. The system's ability to capture shape-independent lip patterns ensures performance is not substantially compromised. These findings demonstrate our biometric system's resilience to viewing angle changes, making it suitable for varied real-world scenarios.

\subsection{Impact of Ambient Lighting}
To assess the impact of ambient lighting on our biometric authentication system, we conducted tests under two distinct lighting conditions with 14 subjects. The first condition involved turning off the overhead light directly above, resulting in an illumination level of approximately 80 ± 10 lux. The second condition utilized fill lighting, set to 500 ± 10 lux, to provide a brighter and more evenly distributed light source.

\begin{figure}[!t]
    \centering
    \includegraphics[width=0.6\linewidth]{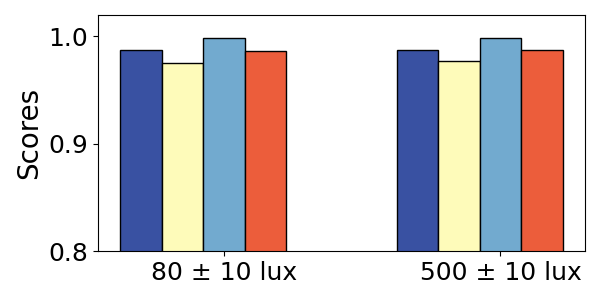}
    \includegraphics[width=0.8\linewidth]{figs/exp/other/legend.png}
    \caption{Evaluation of the impact of ambient lighting.}
    \label{fig:light}
\end{figure}

As demonstrated in Fig. \ref{fig:light}, while variations in ambient lighting did influence image quality, our system demonstrated robustness in both conditions. Although lower illumination with the overhead light slightly reduced image contrast and clarity, it did not significantly impair recognition accuracy. 
This robustness is attributed to our lip print processing, which effectively captures shape-independent lip patterns. Consequently, our system maintains 

\subsection{Impact of Lip wetness and color}

To evaluate the impact of lip wetness and color on our biometric system, we tested various lip conditions with 14 subjects. The evaluation involved applying transparent lip balm, two different lipsticks (``Into You" and ``Mac"), a combination of lip balm and lipstick.
The images were captured under controlled lighting to analyze how these conditions affected the recognition of lip patterns.
\begin{figure}[!t]
    \centering
    \includegraphics[width=0.95\linewidth]{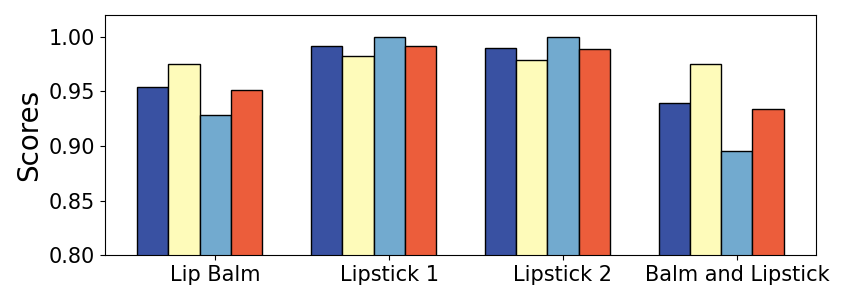}
    \includegraphics[width=0.8\linewidth]{figs/exp/other/legend.png}
    \caption{Evaluation of the impact of lip wetness and color.}
    \label{fig:wet_color}
\end{figure}

As illustrated in Fig. \ref{fig:wet_color}, the results revealed that lip wetness and color have a minor impact on authentication accuracy, precision, recall, and F1 scores, but the effect is relatively minor. Transparent lip balm increased glossiness, which slightly reduced pattern clarity but did not significantly affect overall performance. The colored lipsticks, ``Into You" (color 1) and ``Mac" (color 2), maintained high effectiveness across all metrics. The combination of lip balm and lipstick presented minor additional challenges but still preserved robust performance. 
These findings highlight that while our system is generally robust, certain conditions, such as altered lip boundaries, can impact performance. 

\section{Related Works}
\label{sec:related_work}

\subsection{Biometric Authentication}
Advanced biometric methods are emerging to address the vulnerabilities of traditional features like faces, voiceprints, and fingerprints, such as photoplethysmograms \cite{chen2017your} and 3D geometry \cite{xu2021rface} for face authentication, time-difference-of-arrival dynamic \cite{zhang2016voicelive} for voice authentication.
Other methods include cross-domain speech similarity \cite{shi2020wearid}, physical vibrations of finger touches \cite{xu2020touchpass,yang2021enabling}, and novel biological signals such as bone-conduction breathing \cite{han2023breathsign}, occlusal sounds \cite{xie2022teethpass}, respiratory motion characteristics \cite{liu2020continuous}, 
pulsatile signals from wrist-worn sensors \cite{zhao2021robust}, dental edge biometrics \cite{Jiang2020SmileAuth}, toothprint-induced sonic effects \cite{wang2022toothsonic},
pop noise from close microphone breathing \cite{wang2019voicepop}, 
vocal tract features \cite{lu2020vocallock}, 
heart motion and body asymmetry \cite{cao2023heartprint}, body gestures \cite{shi2017smart}, 
finger gestures \cite{kong2020continuous}, 
hand geometry and behavioral traits \cite{song2017multi}, 
device usage patterns \cite{neal2017mobile}, 
and various biological signals \cite{santos2018ecg,kassab2022ferst}. 
Although lip movement has mostly been used in speech recognition \cite{fan2023mmmic}, recent studies also utilize Doppler profiles of acoustic signals from speaking lips \cite{lu2018lippass}.

\subsection{Lip-based Authentication}
With the limitations of facial and voice authentication regarding privacy and security, lip-based authentication has garnered significant interest. Traditional methods have largely focused on static images, with research demonstrating varying accuracies. For example, machine learning techniques such as the Hough transform for groove extraction~\cite{smacki2010lip, smacki2010lip2}, the Generalized Hough Transform for fragment recognition~\cite{wrobel2013method}, and edge detection methods~\cite{jain2018cheiloscopic} have shown promising results but often lack the ability to handle dynamic lip movements effectively. In contrast, deep learning methods, while still emerging, have shown improved accuracy with VGG16 and VGG19~\cite{farrukh2023lip}, and other architectures like the convolutional network~\cite{zhou2023lip}. However, these approaches typically focus on static image features and have yet to fully address the complexities of dynamic lip movements.

\subsection{Dynamic lip-based authentication}
Studies on dynamic lip-based authentication have explored various approaches, though many still focus on static features~\cite{chen2006video}. 
Other studies combined edge and LBP features with DTW alignment and AdaBoost classification, but still primarily relied on static features~\cite{sayo2011biometrics}. 
Yang et al. developed a deep learning model for lip-based authentication but lacks interpretable insights into the fundamental characteristics and patterns of lip articulation during speech~\cite{Yang2021Preventing} .
Researchers also attempted to incorporate dynamic features by tracking lip contours and extracting motion features, but their methods, such as GMM and HMM, resulted in lower accuracies due to limited consideration of lip movement dynamics~\cite{cetingul2006discriminative}.

\section{Conclusion}
\label{sec:conclusion}

In this paper, we proposed a robust biometric authentication system based on dynamic lip movements. We developed a comprehensive understanding of shape-independent lip patterns and dynamic features, building a feature hierarchy for detailed representations of lip dynamics. Our system addresses a critical limitation in real-world applications, i.e., the natural alternation between static and dynamic lip states during daily activities. Unlike traditional methods that only work in restricted conditions, our approach effectively handles both scenarios: when users are in a natural, non-speaking state and when they are actively speaking. We evaluated the system's effectiveness across these varying scenarios, demonstrating superior performance regardless of lip status. Additionally, our evaluation against multiple attack models highlights the system's resilience and security, proving the advantages of our feature hierarchy approach for reliable biometric authentication in real-world applications.

\bibliographystyle{unsrt}
% \bibliography{bib/dynamiclip}

\begin{IEEEbiography}
[{\includegraphics[width=1in,height=1.25in,clip,keepaspectratio]{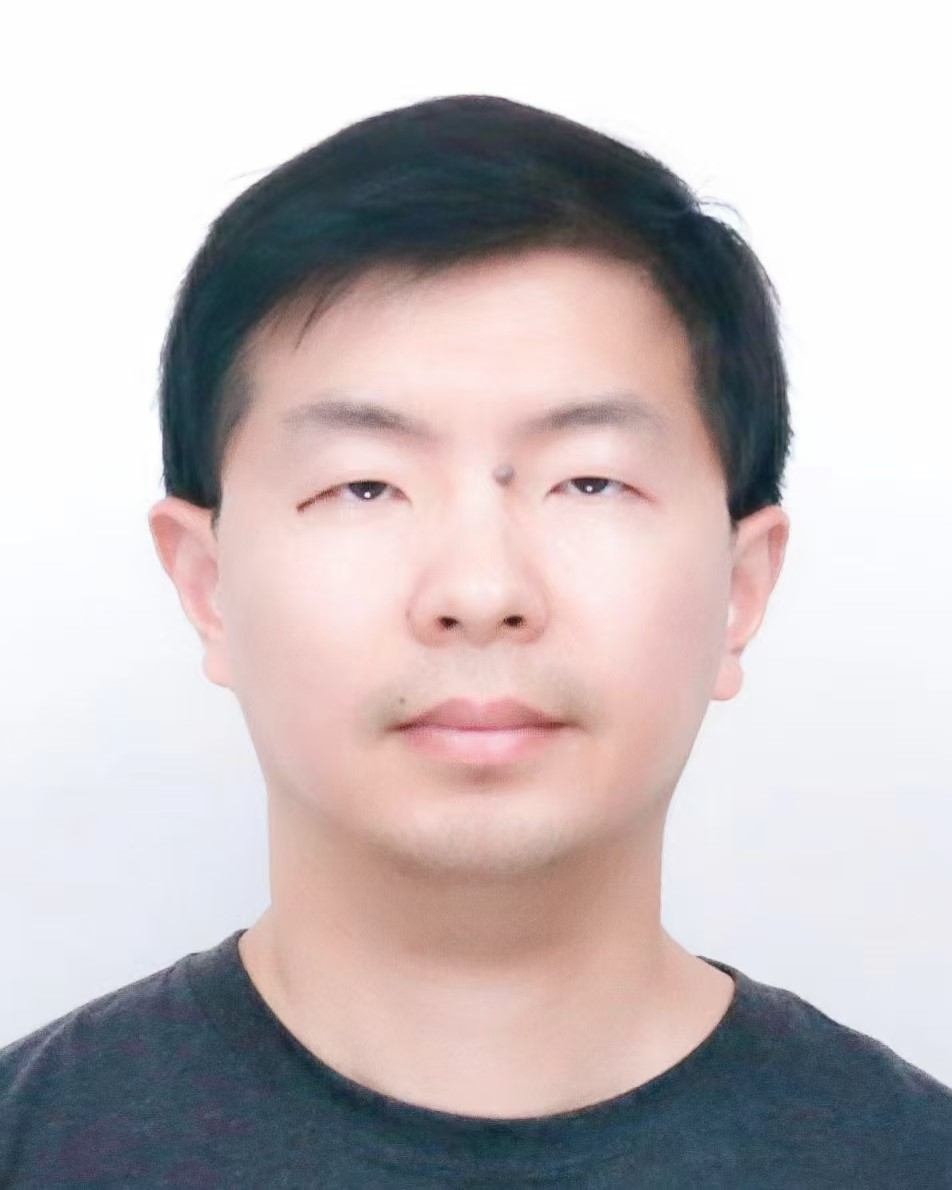}}]
{Huashan Chen} is now a Young Associate Professor at the Institute of Information Engineering, Chinese Academy of Sciences. He received the B.S. and M.S. degrees from Shandong University, and the Institute of Information Engineering, Chinese Academy of Sciences, respectively. He received the Ph.D. degree in Computer Science from The University of Texas at San Antonio in 2021. His research interests include cybersecurity metrics and biometric authentication.
\end{IEEEbiography}

\vspace{-4\baselineskip}

\begin{IEEEbiography}
[{\includegraphics[width=1in,height=1.25in,clip,keepaspectratio]{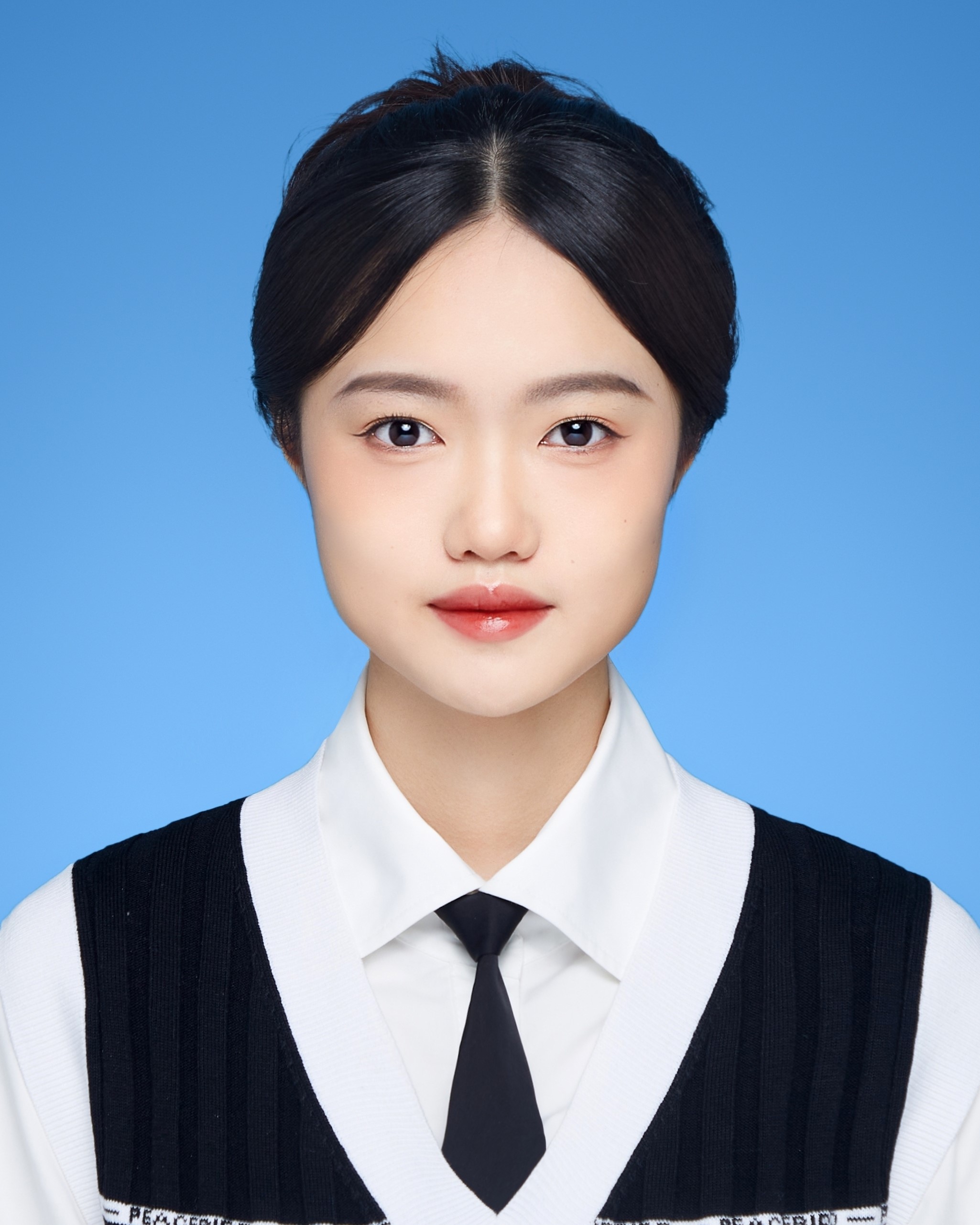}}]
{Yifan Xu} received the BSc degree from Software College of Northeastern University, Liaoning, China, in 2023. She is currently working toward the Ph.D. degree with Institute of Information Engineering, Chinese Academy of Sciences, Beijing, China. Her research of interests are in the areas of cyberspace security and biometric authentication.
\end{IEEEbiography}

\vspace{-3\baselineskip}

\begin{IEEEbiography}
[{\includegraphics[width=1in,height=1.25in,clip,keepaspectratio]{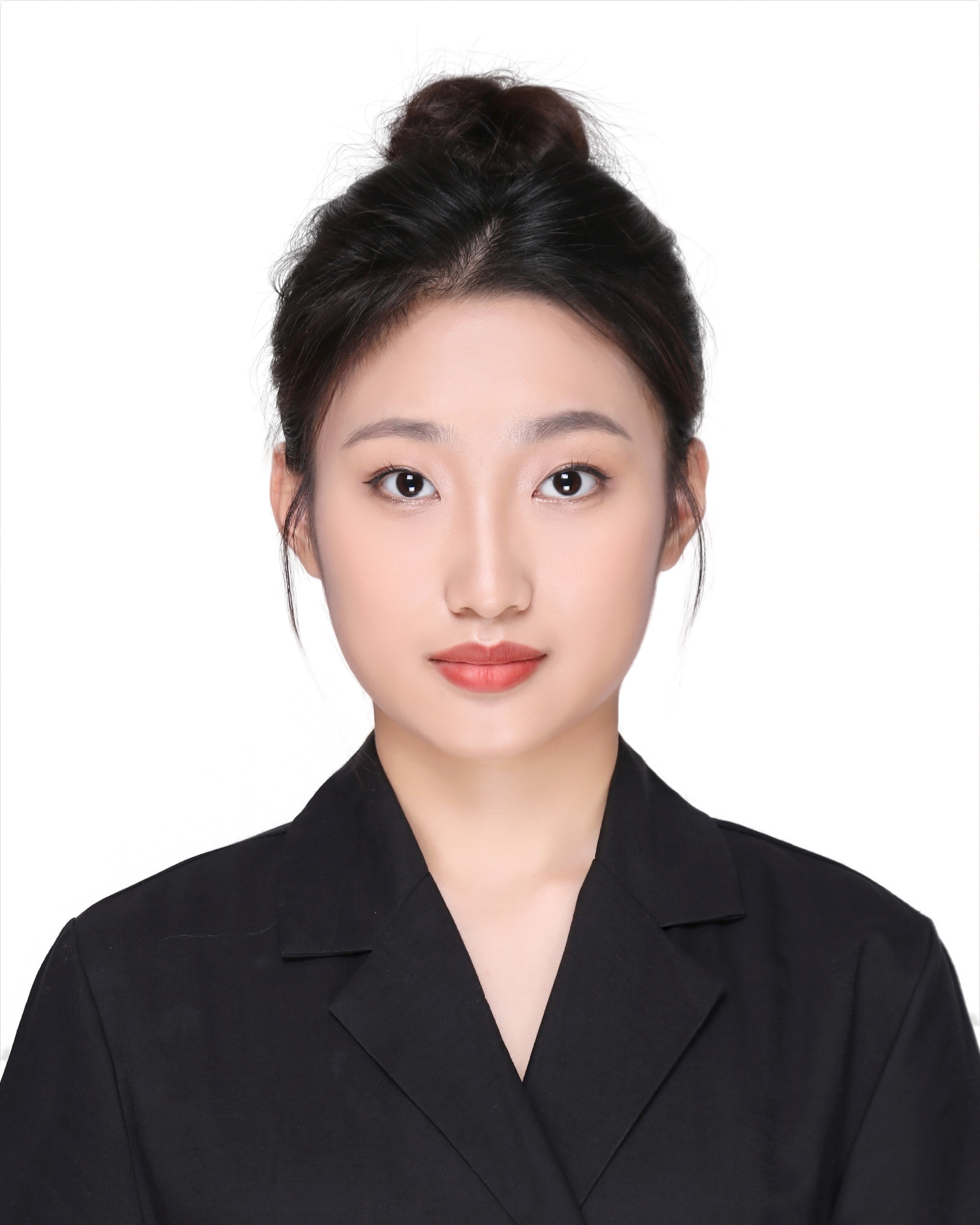}}]
{Yue Feng} received the BSc degree from the School of Mechanical Electronic and Information Engineering, China University of Mining and Technology (Beijing), in 2023. She is currently working toward the Ph.D. degree with the Institute of Information Engineering, Chinese Academy of Sciences. Her research of interests are in the areas of cyberspace security, audio/visual signal processing, biometric authentication.
\end{IEEEbiography}

\vspace{-3\baselineskip}

\begin{IEEEbiography}
[{\includegraphics[width=1in,height=1.25in,clip,keepaspectratio]{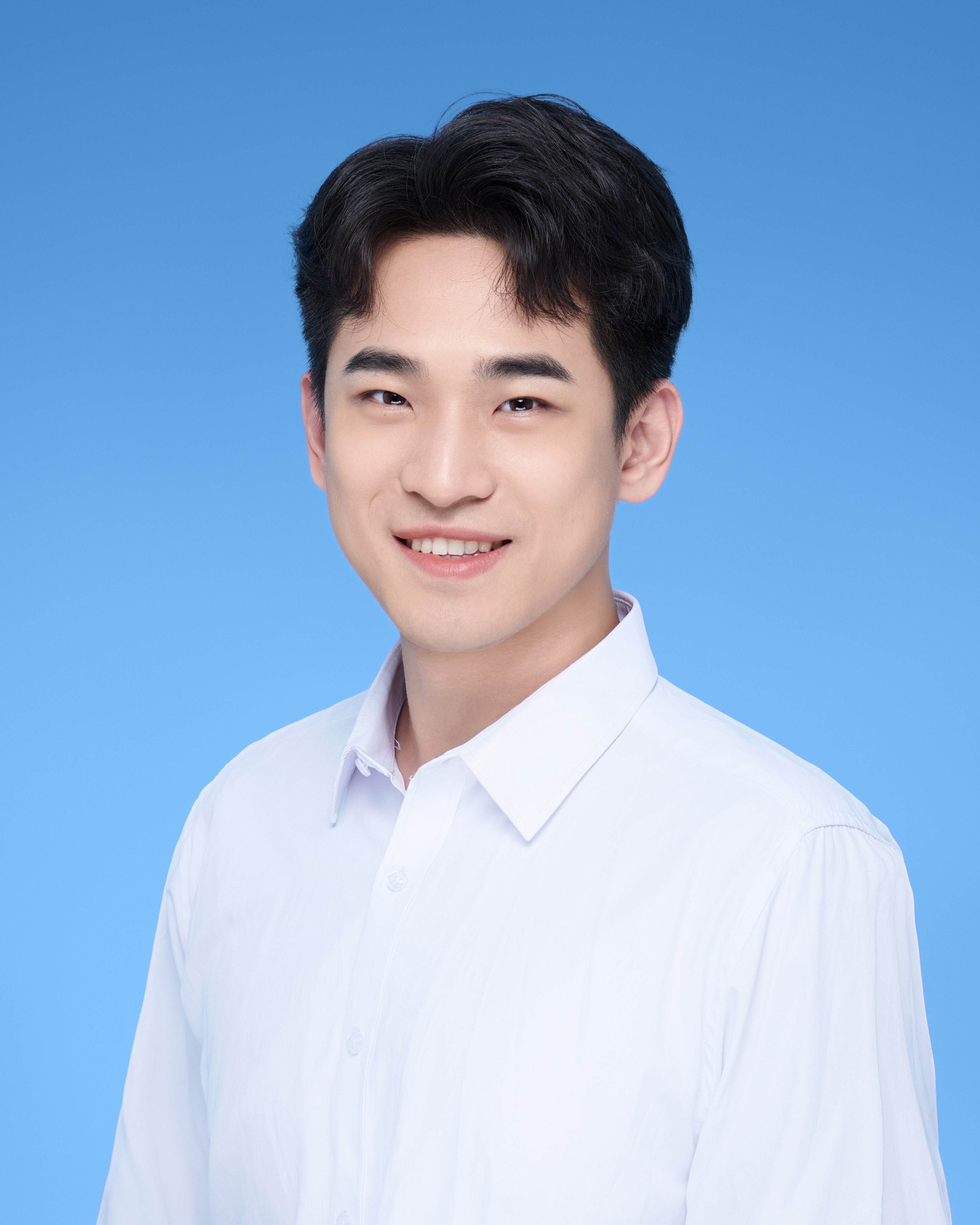}}]
{Ming Jian} received the B.S. degree in Information Technology from St.John's University in 2019. He is currently an engineer at the Institute of Information Engineering, Chinese Academy of Sciences. His research interests include cybersecurity.  
\end{IEEEbiography}

\vspace{-3\baselineskip}

\begin{IEEEbiography}
[{\includegraphics[width=1in,height=1.25in,clip,keepaspectratio]{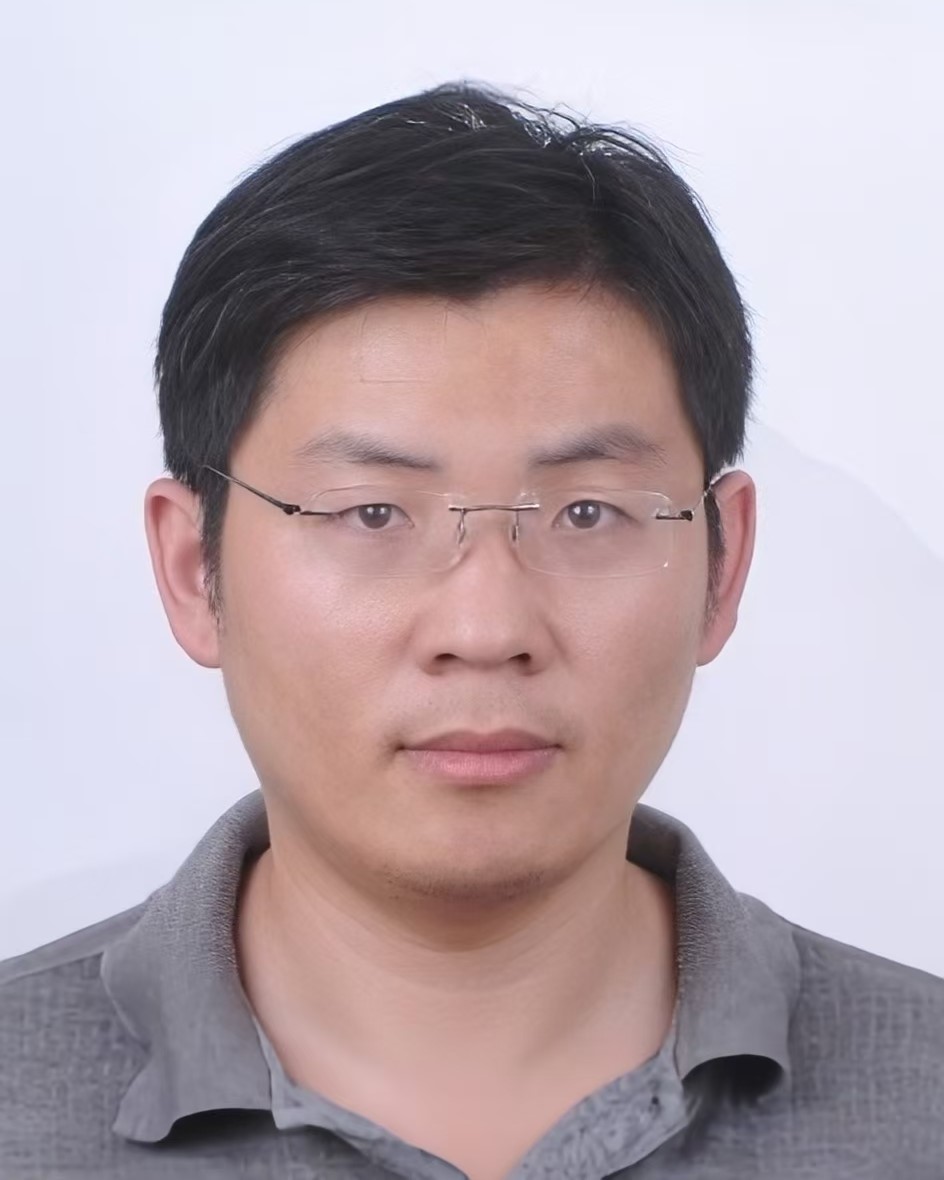}}]
{Feng Liu} is a Professor and Ph.D. supervisor at the Institute of Information Engineering, Chinese Academy of Sciences. He is also a professor in the School of Cyber Security, University of Chinese Academy of Sciences. He received his B.S. degree in 2003 from Shandong University and the Ph.D. degree in 2009 from Institute of Software, Chinese Academy of Sciences. His research interests include strategic and economic aspects of information security, system security, visual security and cryptography. He co-initiated the International Conference on Science of Cyber Security (SciSec) and is serving as its Steering Committee Chair. He serves as the Editor-in-Chief of the International Journal of Digital Crime and Forensics (IJDCF).  
\end{IEEEbiography}

\vspace{-3\baselineskip}

\begin{IEEEbiography}[{\includegraphics[width=1in,height=1.25in,clip,keepaspectratio]{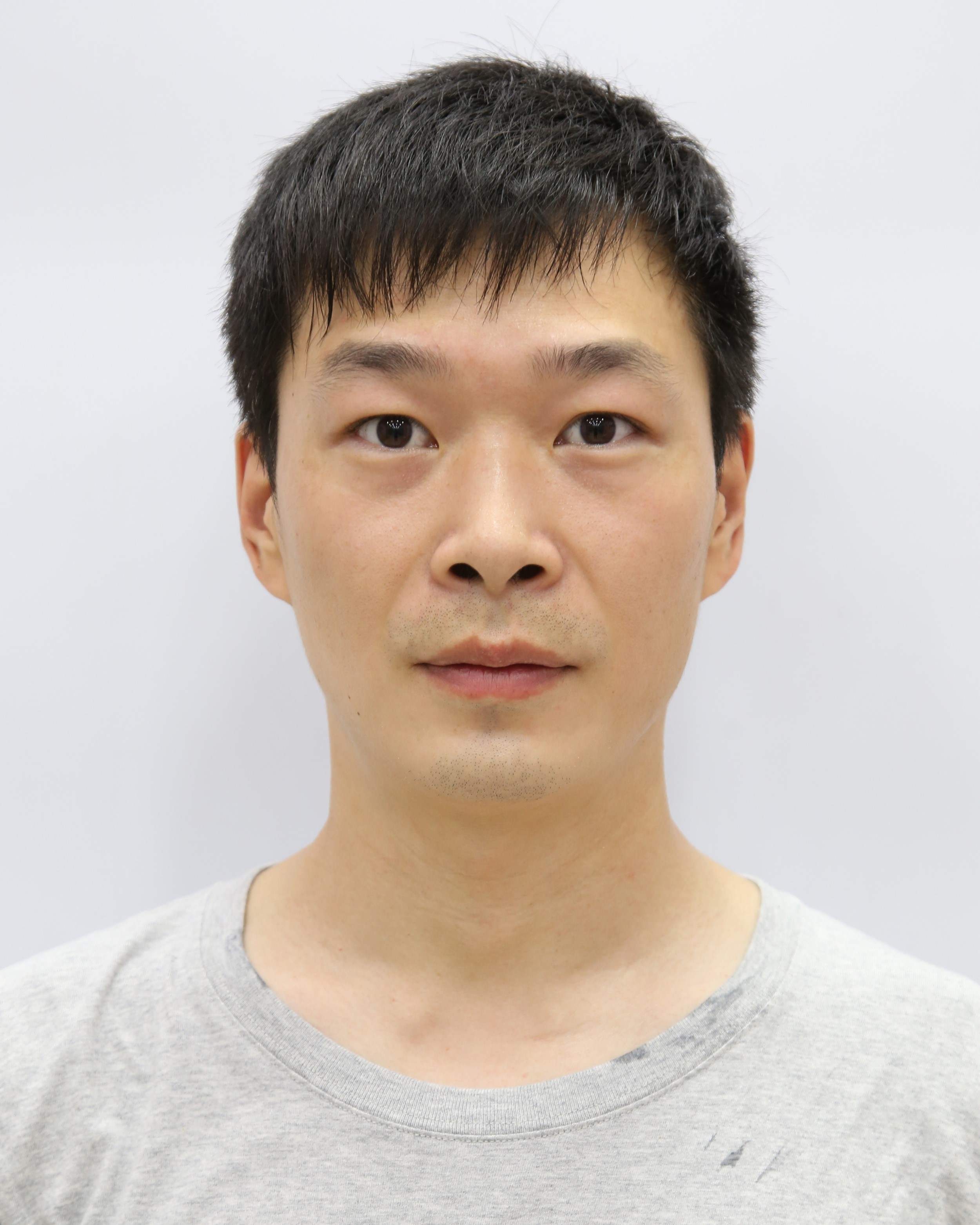}}]
{Pengfei Hu} is a Professor in the School of Computer Science and Technology at Shandong University. He received Ph.D. in Computer Science from UC Davis. His research interests are in the areas of cyber security, data privacy, and mobile computing. He has published more than 60 papers in premier conferences and journals on these topics. He served as TPC for numerous prestigious conferences, and associate editors for IEEE TWC and IEEE IoTJ.
\end{IEEEbiography}

\vspace{-1\baselineskip}

\begin{IEEEbiography}
[{\includegraphics[width=1in,height=1.25in,clip,keepaspectratio]{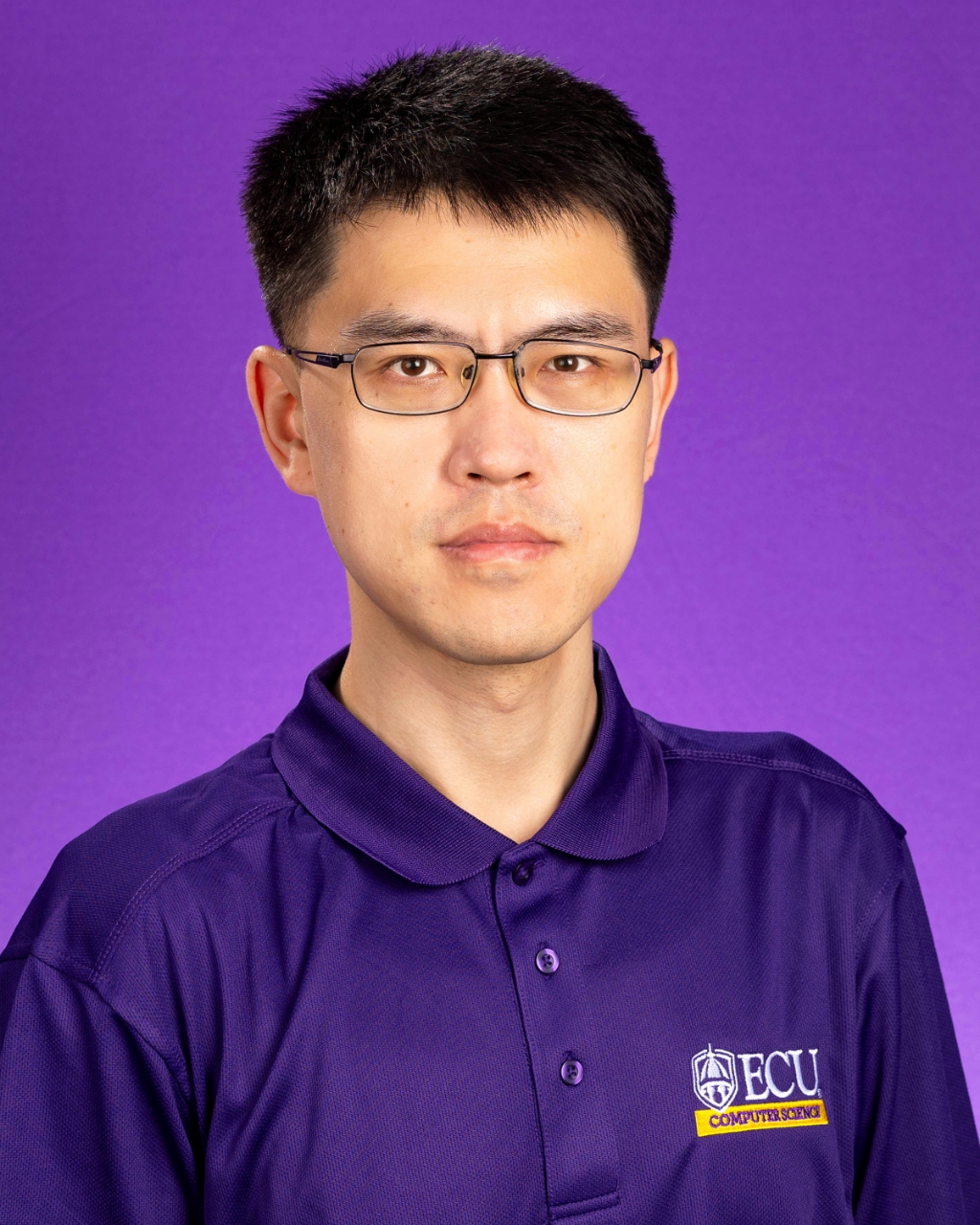}}]
{Kebin Peng} is an Assistant Professor of Computer Science at East Carolina University. His research interests include deep learning, large language models for software performance testing, code generation, and computer vision, with a focus on autonomous driving, detection, understanding, and reconstruction. He earned his Ph.D. in Computer Science from The University of Texas at San Antonio and has industry experience as a Senior Software Engineer at MathWorks, where he contributed to the Embedded Coder product team. He has published his research in top-tier conferences, including CVPR, ICPR, and others, showcasing his contributions to cutting-edge advancements in the field.
\end{IEEEbiography}

\vspace{-1\baselineskip}

\begin{IEEEbiography}
[{\includegraphics[width=1in,height=1.25in,clip,keepaspectratio]{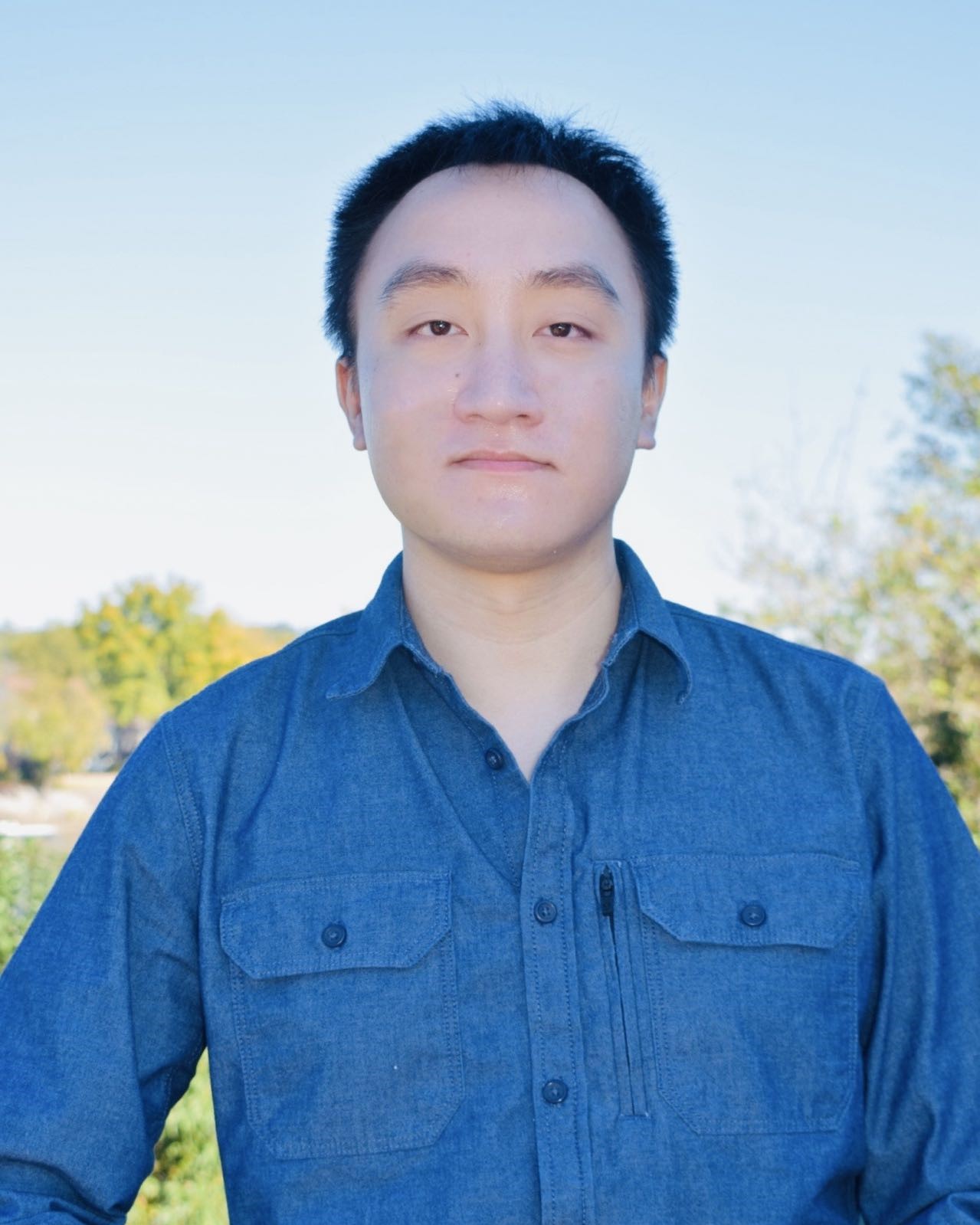}}]
{Sen He} is an Assistant Professor in the Software Engineering Program, Department of Systems and Industrial Engineering at the University of Arizona. Sen He received his Ph.D. from The University of Texas at San Antonio in July 2022. Sen He’s research interests include the areas of Cloud Computing, Software/Performance Engineering (Formal Methods, LLM for SE), and Computer Vision. His works have been published in flagship conferences, including ESEC/FSE, MICRO, ASE, ECCV, and EuroSys. And one of his main works has received the ACM SIGSOFT Distinguished Paper Award.
\end{IEEEbiography}

\vspace{-1\baselineskip}

\begin{IEEEbiography}
[{\includegraphics[width=1in,height=1.25in,clip,keepaspectratio]{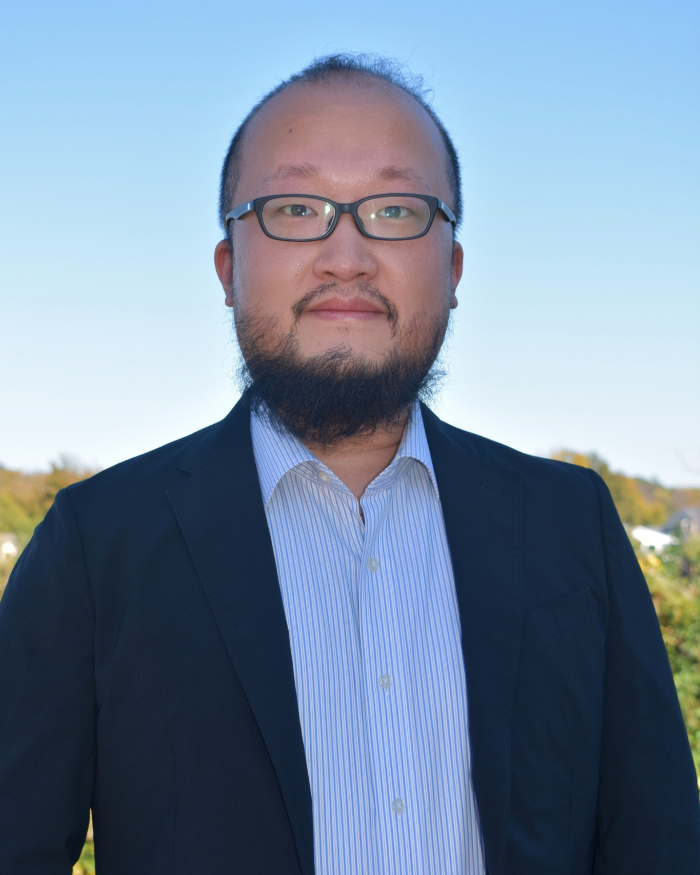}}]
{Zi Wang} is an Assistant Professor of Computer Science in the School of Computer and Cyber Sciences at Augusta University. Zi Wang received his Ph.D. from the Department of Computer Science at Florida State University. His research interests span Mobile Computing, Human-Computer Interaction and AIoT. He published papers in premier venues such as ACM Ubicomp, ACM MobiCom, and ACM CHI. His current projects include earables sensing, human biometrics, user authentication, and mobile sensing for human-computer interaction applications. 
\end{IEEEbiography}

\vfill
\end{document}